\documentclass[pdflatex,sn-basic]{sn-jnl}

\usepackage{graphicx}%
\usepackage{multirow}%
\usepackage{amsmath,amssymb,amsfonts}%
\usepackage{amsthm}%
\usepackage{mathrsfs}%
\usepackage[title]{appendix}%
\usepackage{xcolor}%
\usepackage{textcomp}%
\usepackage{manyfoot}%
\usepackage{booktabs}%
\usepackage{newunicodechar}
\newunicodechar{≈}{\approx}
\usepackage{subcaption}
\usepackage{placeins} 
\usepackage{algorithm}%
\usepackage{algorithmicx}%
\usepackage{algpseudocode}%
\usepackage{comment}
\usepackage{listings}%
\usepackage{float}
\usepackage{booktabs}
\usepackage{array}
\usepackage{tabularx}
\usepackage{amsmath}

\theoremstyle{thmstyleone}%
\theoremstyle{thmstyletwo}%

\theoremstyle{thmstylethree}%

\begin{document}

\title[Article Title]{\textcolor{red}{GeMi}: A \textcolor{red}{\textbf{G}}raph-bas\textcolor{red}{\textbf{e}}d, \textcolor{red}{\textbf{M}}ult\textcolor{red}{\textbf{i}}modal Recommendation System for Narrative Scroll Paintings}


\author*[1,2]{\fnm{Haimonti} \sur{Dutta}}\email{haimonti@buffalo.edu}


\author[3]{\fnm{Pruthvi} \sur{Moluguri}}\email{pmolugur@buffalo.edu}

\author[1]{\fnm{Jin} \sur{Dai}}\email{jindai@buffalo.edu}

\author[4]{\fnm{Saurabh Amarnath} \sur{Mahindre}}\email{smahindr@buffalo.edu}

\affil[1]{\orgdiv{Department of Management Science and Systems}, \orgname{The State University of New York}, \orgaddress{\street{Jacobs Management Center}, \city{Buffalo}, \postcode{14260}, \state{NY}, \country{USA}}}

\affil[2]{\orgdiv{Institute for Artificial Intelligence and Data Science}, \orgname{The State University of New York}, \orgaddress{\street{Lockwood Hall}, \city{Buffalo}, \postcode{14260}, \state{NY}, \country{USA}}}

\affil[3]{\orgdiv{Department of Computer Science and Engineering}, \orgname{The State University of New York}, \orgaddress{\street{Davis Hall}, \city{Buffalo}, \postcode{14260}, \state{NY}, \country{USA}}}

\affil[4]{\orgdiv{Technology, Information, Internet}, \orgname{eBay\footnote{This work was done prior to the author joining his current affiliation.}}, \orgaddress{\street{Embassy Tech Village}, \city{Bengaluru}, \postcode{560103}, \state{Karnataka}, \country{India}}}


\abstract{Recommendation Systems are effective in managing the ever-increasing amount of multimodal data available today and help users discover interesting new items. These systems can handle various media types such as images, text, audio, and video data, and this has made it possible to handle content-based recommendation utilizing features extracted from items while also incorporating user preferences. Graph Neural Network (GNN)-based recommendation systems are a special class of recommendation systems that can handle relationships between items and users, making them particularly attractive for content-based recommendations. Their popularity also stems from the fact that they use advanced machine learning techniques, such as deep learning on graph-structured data, to exploit user-to-item interactions. The nodes in the graph can access higher-order neighbor information along with state-of-the-art vision-language models for processing multimodal content, and there are well-designed algorithms for embedding, message passing, and propagation. In this work, we present the design of a GNN-based recommendation system on a novel data set collected from field research. Designed for an endangered \emph{performing} art form, the recommendation system uses multimodal content (text and image data) to suggest similar paintings for viewing and purchase. To the best of our knowledge, there is no recommendation system designed for narrative scroll paintings -- our work therefore serves several purposes, including art conservation, a data storage system for endangered art objects, and a state-of-the-art recommendation system that leverages both the novel characteristics of the data and preferences of the user population interested in narrative scroll paintings. }




\maketitle

\section{Introduction}
\label{intro}

Recommendation Systems (RS) (\cite{Deldjao_20a}, \cite{zhou_23}) help organize a large number of items (such as books, music, appliances) available to users. They find items of interest by predicting the preference for candidate items and recommending the most pertinent ones using historical data.
Data-driven RS falls into three main categories: collaborative filtering (CF), Content-based Filtering (CBF), and Hybrid. In CF models, the emphasis is on a community of users,  their behavioral patterns and preferences.
The CBF methods depend on the user's behavioral data along with information about the items, such as words in a text document, color in images, and audio content. Some recommendation systems are known to combine collaborative filtering methods with content-based methods for added benefits. All three categories of recommendation systems use multimodal data (images, text, audio, video) extensively. 

Conventional methods for integrating data from different modalities in RS include concatenation or summation (\cite{He_16a, Liu_17a}). However, with the advent of deep learning techniques with non-linear modeling capabilities, each modality can be represented separately and combined subsequently to generate a suitable representation. 
The recent increase in interest in graph-based recommendation systems has inspired a new line of work that explores higher-order interactions between users and items by leveraging the power of graph neural networks (\cite{He_20a, Wu_23a, Zhang_21a, Zhang_22a, Zhou_23a, Zhou_23b, Wei_19a, Wei_20a, Wang_23a,Zhou_23b}). For example, MMGCN (\cite{Wei_19a}) exploits user-item interactions to guide representation learning in each modality by constructing graph convolutional networks with nodes having topological features and aggregating features of neighbors. GRCN (\cite{Wei_20a}) refines this technique by designing a graph refining layer that is designed to identify noisy edges with high confidence in user preference. However, user and item representations learned by graph aggregations are often inherently noisy, since user-item interactions are sparse. These systems also suffer from cold start problems, where there is not enough data (such as clicks, ratings, or purchases) to learn preferences.


One solution to the user-item sparsity problem is to reinforce node (users and items) features. However, the noise present in the nodes often makes it difficult to extract useful information for recommendations. Noise in item features (such as blurry images, errors in spelling in text, damaged soundtracks, or misleading descriptions) causes uncertainty in representation. The item features may also contain irrelevant information, such as background color or distracting objects in the image. In multimodal settings, complementary information from different modalities is leveraged to deal with sparsity, but noisy features can lead to undesirable results. If certain modalities are missing or absent, the multimodal system must accommodate this. To address these challenges in multimodal learning, Variational Auto Encoders (VAEs) have become popular (\cite{Yi_22a,Liang_18a,Ma_19a,Li_17a}) in which modality-specific encoders learn a Gaussian variable parameterized by a mean and variance vector for each node. The mean vector is modeled to represent semantic information, and the variance vector captures the uncertainty of the corresponding modality. 

In this paper, we describe a multimodal content-based recommendation system augmented with user preference information, that is designed for a unique, sociocultural application to conserve and study an endangered performing art from eastern India. \emph{Singing painters} are a group of itinerant storytellers who paint scrolls and use a song or poem to explain the artwork. The art is usually inspired by epics, regional folktales, and events that transpire in the day-to-day lives of the painters (\cite{McCutchion_99a}). Songs or poems are typically learned by word-of-mouth. Sporadic, ad hoc efforts have been made to store samples of artwork or song texts in non-digitized formats, but to the best of our knowledge, a meticulous study of this multimodal data from images and text, or the design of a recommendation system to organize it, has never been undertaken. In recent work (\cite{Dutta_24a}), 
we have prepared a custom dataset of digitized images, aligning lyrics of songs with images, and documenting, to a good extent, regional variations in style, theme, color choice, and other relevant details. We use an enhanced version of this dataset\footnote{More data is available from subsequent field studies in other districts of the state.}  to design the recommendation engine, which can suggest artwork similar to what appreciative users like. 
In doing so, it was found that while there have been many techniques for the construction of multimodal recommendation systems for known benchmark datasets (such as Amazon (\cite{Ni_19a}), MovieLens (\cite{Harper_15a}) or Netflix (\cite{Bennett_07a})), the application of these to a novel data set can quickly become tedious if careful analysis and principled experimentation are not adhered to. We report our findings on the design of the multimodal recommendation system, GeMi, carefully benchmarking it with state-of-the-art systems found in the literature. The dataset used in this research is available from: \url{https://github.com/Haimonti/Folk-Art-Recommendation-System}.

\noindent The primary contributions of this paper are:
\begin{itemize}
    \item A unique, multimodal dataset collected from field surveys and curated for research
    \item The GeMi Recommendation System - A content-based recommendation using state-of-the-art vision-language models with sophisticated multimodal fusion, graph structure learning, and incorporating user preferences through collaborative filtering style plug-and-play modules.
\end{itemize}
This paper is organized as follows: Section~\ref{related} presents Related Work; the data collection procedure along with non-digitized sources are described in Section~\ref{data}; the recommendation system is presented in Section~\ref{arch}; empirical results comparing the performance of GeMi to state-of-the-art transductive and inductive homogeneous and heterogeneous graphs are presented in Section~\ref{EmpResults}; discussion of enhancements possible is proposed in Section~\ref{disc}, and Section~\ref{conc} concludes the paper.

\section{Related Work}
\label{related}

In this section, we present prior work on art recommendation and a 
brief history of narrative scroll paintings. Appendix~\ref{GnnReco} provides a review of the use of multimodal data, Variational Auto Encoders (VAEs), and Large Language Models (LLMs) to augment Graph Neural Networks (GNNs) for building recommendation systems. 

\subsection{Art Recommendation}

Museums have recognized the need to provide visitors with personalized experiences when visiting collections and have adopted recommendation systems to meet this requirement. For example, a visitor's profile might store the names of their favorite artists or their painting techniques, extracted from short textual descriptions associated with the artworks (\cite{Wang_08, Wang_09}). The user profile is then matched to the features of the items to provide personalized suggestions. The Cultural Heritage Information Personalization (CHIP) project (\cite{Aroyo_07}) built a recommendation engine for the Rijks museum in Netherlands.
In the Folksonomy-based Item Recommender syStem (FIRSt) (\cite{Semeraro_12}), user-generated content from social tagging is integrated into a content-based recommender, allowing users to express item preferences by entering a numerical rating in addition to annotating items with free tags. In the Personal Experience with Active Cultural Heritage (PEACH) project (\cite{Stock_07, Messina_19a, Kuflik_11a}), a PDA-based museum tour application is developed, which is location-aware and whose content is adapted to the needs of the visitor. Approaches that make suggestions to museum visitors based on their history within the physical environment and textual information associated with each item in their history have become fairly popular (\cite{Bohnert_09a, Grieser_11a}). Yilma et al. (\cite{Yilma_23a}) focus on efficiently capturing the latent semantic relationships of visual art for personalized recommendation. They show that textual features are as useful as visual ones, but a fusion of both captures most of the latent semantic information. Kuflik et al. (\cite{Kuflik_11a}) showed the benefits of using graph-based recommendation. An anti-recommendation system, entitled ``Art I don't like" (\cite{Frost_19}), suggests artwork dissimilar to the ones users like, thereby suggesting that removing access to opposing viewpoints can lead to \textit{filter bubbles} (\cite{Pariser_11a}) and intellectual isolation. 
However, to the best of our knowledge, an art recommendation system solely designed for narrative scroll paintings does not exist to date.


\subsection{A brief history of the dissemination of narrative scroll paintings in South Asia}
\label{secA1}

Narrative scroll paintings are a popular form of \emph{performative} art in Eastern India. It incorporates several modes of communication—visual messages, oral traditions, and music — which blend to describe social transformations, stories of migrations, and socio-political and religious reflections (\cite{Korom_book, Bajpai_15a}). 
Although the exact date of origin of this profession is not known, ancient lore and traditions indicate that it dates back to the 5th-6th century (possibly even earlier) (\cite{Mair_19a}). 

These paintings from eastern India have spread to other parts of the country, including the western and southern regions (such as Deccani picture showmen, the ``Paithan" paintings of Chitrakathis in Maharashtra, the ``Garoda" showmen of Gujarat, the narrative paintings of Telengana and the Bhopas of Rajasthan) (\cite{Jain_98a}). Furthermore, confronted by foreign invasions in the east, many scroll paintings have found their way to neighboring countries such as Nepal, Tibet, and other regions along the Silk Road. These traditions were also used to spread religious learning and made their way into China. 
It is also known that the technique of picture recitation made its way to the western heartland of the Turkish-speaking world by the end of the sixteenth century and possibly much earlier. Evliyā Efendi (ca. 1611–1660) provides a fascinating description of a group known as ``Painter Fortunetellers" in his account of life in Istanbul. It must also be noted that in Indonesia, dramatic performances were well known as \emph{wayang} - a transformative realization used for religious instruction. Wayang means ``shadow," and thus picture scrolls, puppets, shadows, and dancers are all technically wayang. Quoting Mair (\cite{Mair_19a}), ``Indian inputs were essential for the formation of (Chinese) pien, and it is even more obvious that wayang owes an enormous debt to Indian storytelling and theater." Unfortunately, very little of this art form has been documented or digitized.



\section{Data}
\label{data}

 In recent work, the first author undertook extensive field surveys in eastern India\footnote{This involved surveying villages in the districts of Birbhum, Bankura and Purulia in West Bengal.} which has allowed digitization of the scrolls, video, and audio recordings of performances, followed by transcription, transliteration, and translation (original songs are in Bengali) of lyrics of songs. The data from this field study was used to design the first version of the content-based recommendation system (\cite{Dutta_24a}) for art-lovers and curators. A second field survey was conducted a year later, helping to augment the data in the repository. It must be noted that while the images and text collected from the first survey were well aligned, those from the second were not. The missing data (either the text of the song was not available or there was no painting) led to several challenges in the alignment of multimodal data. The data collected remain the copyright of the artists and song creators, but the Institutional Repository located at the State University of New York at Buffalo has obtained permission and consent to archive and store it. 


\subsection{Existing non-digitized sources of data}
\label{ExistingSrcs}
\begin{itemize}
\item \textbf{Text:} There is no known database of songs that can be leveraged for the design of a recommendation system. Therefore, the process of collecting songs and their lyrics had to be done manually through field studies. However, we are aware of several sources where some preliminary attempts have been made to collect the songs: (1) Books (\cite{McCutchion_99a}, \cite{Dutta_39}, \cite{Dimock_63}) and theses (\cite{Yazijian_07a} (2) Personal collections, Dr Korom (\cite{Korom_book}) – these are available on tapes and currently not digitized. 

\item \textbf{Images:} Existing non-digitized narrative scroll paintings are found in the collections of the British Museum, the Museum of International Folk Art (Santa Fe, New Mexico), Victoria and Albert Museum (UK), and the Gurusaday Museum (Kolkata, India). 
However, it must be noted that the process of alignment of images with songs is non-trivial in most cases, since sporadic conservation efforts have varying objectives and may not contain all the relevant data necessary for our study. 

\end{itemize}

 \begin{figure}[!t]
\centering
\includegraphics[width=0.58\textwidth,height=0.66\textwidth]{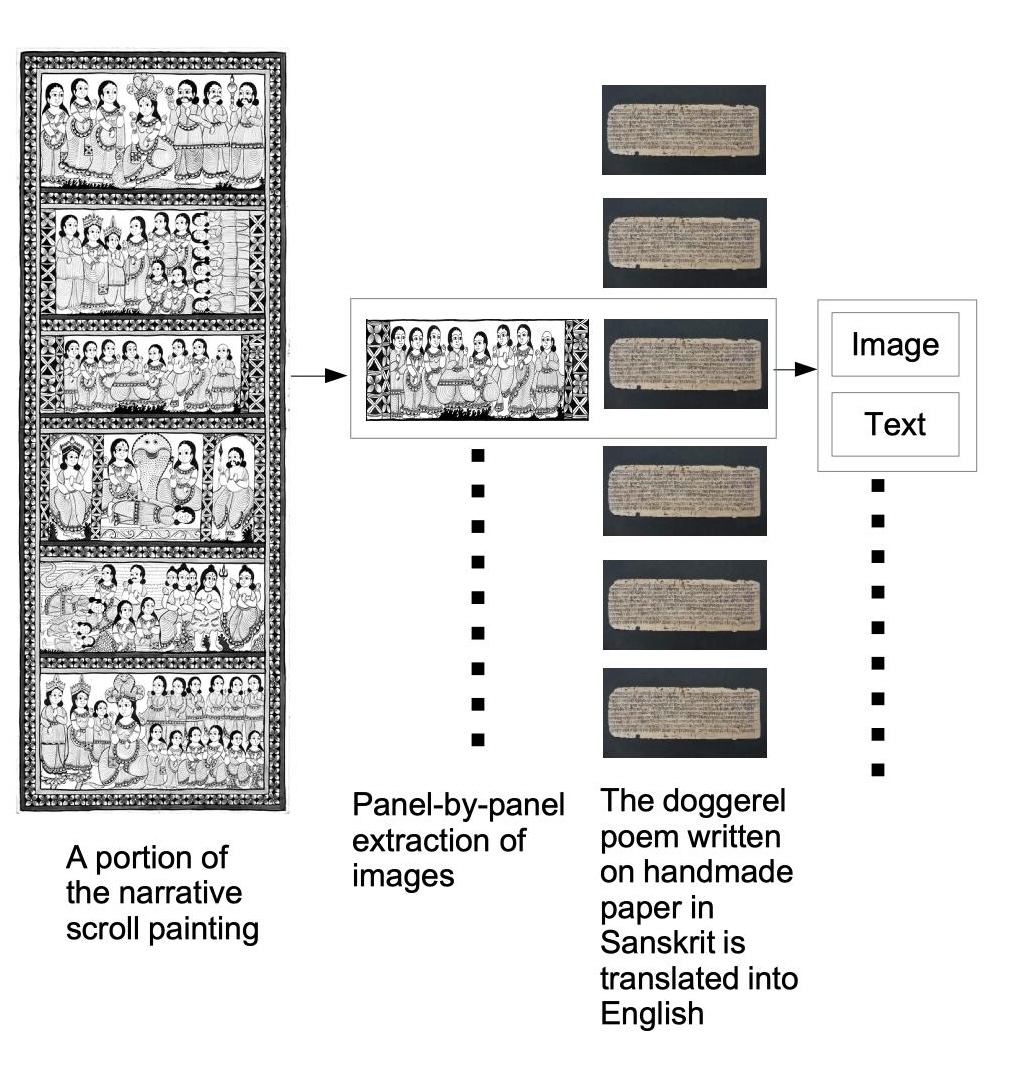}
\caption{An illustration of a panel-by-panel extraction of images from a narrative scroll painting and alignment with the text of an ancient doggerel poem written on handmade paper using black ink. Many parts of the original manuscript of the poem are missing or lost.}
\label{fig:panel}
\end{figure}

\section{The Recommendation System}
\label{arch}

\begin{figure}[!t]
\centering
\includegraphics[width=0.88\textwidth,height=1.2\textwidth]{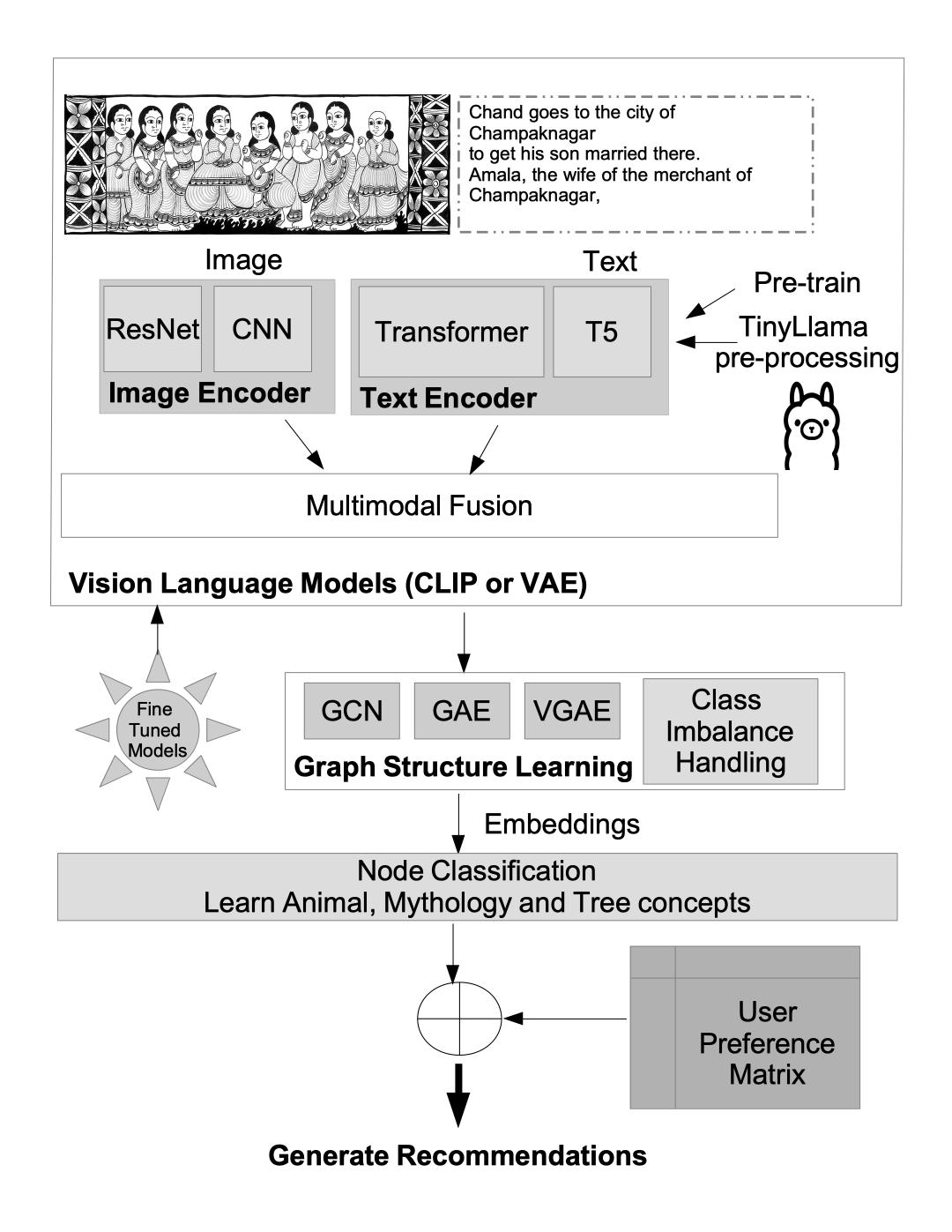}
\caption{The GeMi recommendation system. Fine-tuned Vision Language models are used to extract features, learn latent graph structures, and embeddings using graph convolutions. User-Panel preference matrices are used to augment recommendations.}
\label{fig:recsys}
\end{figure}

\noindent Figure~\ref{fig:panel} provides an illustration of the multimodal data extraction procedure and Figure~\ref{fig:recsys} represents the overall architecture of the recommendation system. It has the following components: (a) Fine-tuned vision language models for multimodal feature extraction (b) Graph Structure Learning (c) Node Classification and, (d) downstream content-based recommendation. The following sections describe them in more detail. \\ 

\noindent \textbf{Preliminaries} 
Our basic premise is that a recommendation of panels and songs should reflect the concept(s) a user is interested in. We study the user-panel preference matrices to develop an algorithm for recommendation. Let $\mathcal{U}, \mathcal{P}$ denote the set of users and panels, respectively. Each user $u_i \in \mathcal{U}$ is associated with a set of panels $p_j \in \mathcal{P}, i \in {1,\cdots, |\mathcal{U}|}; j \in {1,\cdots, |\mathcal{P}|}$ which s/he likes. Figure~\ref{fig:panel} shows a schematic of the panel-by-panel image extraction. The preference score $y_{ij}=1$, if the user $u_i$ likes panel $p_j$ and 0 otherwise. Each panel $p_j$ has a multimodal representation i.e. $p_j \in R^{D_m}$ where $D_m$ represents the dimensionality of modality $m \in \mathcal{M}$ and $\mathcal{M}$ is the set of modalities. The multimodal recommendation seeks to accurately find panels a user might like, given his/her known preferences. 

In the following subsections, we describe each component of the GeMi recommendation system.

\subsection{Fine-tuned vision language models for multimodal feature extraction}
\label{SR}


We describe a multimodal feature extraction framework that transforms image and text associated with narrative scroll paintings into semantically structured, high-dimensional representations suitable for downstream graph-based recommendation.

Given the symbolic density, narrative continuity, and cultural specificity inherent in them, unimodal representations are insufficient to capture the full semantic spectrum of the data. Instead, we adopt a modular multimodal encoding strategy that integrates complementary inductive biases from language modeling, cross-modal contrastive learning, and probabilistic latent-variable modeling.

Formally, let each panel be represented by a tuple
$\mathcal{P}_i = (x_i^{\mathrm{img}}, x_i^{\mathrm{text}}),$
where $x_i^{\mathrm{img}} \in \mathcal{X}_{\mathrm{img}}$ denotes the visual depiction of the panel and
$x_i^{\mathrm{text}} \in \mathcal{X}_{\mathrm{text}}$ denotes its associated textual description.
Our objective is to construct a family of embedding functions
\begin{equation}
\Phi^{(k)} : \mathcal{X}_{\mathrm{img}} \times \mathcal{X}_{\mathrm{text}} \rightarrow \mathbb{R}^{d_k},
\end{equation}
each capturing distinct semantic views of the same panel content, which are later integrated through graph-based propagation. To this end, we employ three conceptually distinct yet complementary representation mechanisms:
\begin{enumerate}
    \item A Large Language Model (LLM) (LLaMA, \cite{Touvron_23}) for semantic abstraction and canonicalization of textual descriptions.
    \item A sigmoid-Contrastive Language Image Pretraining model (Sig-CLIP, \cite{radford_21a}) for joint image-text alignment,
    \item A multimodal variational autoencoder (VAE) for probabilistic fusion of visual and linguistic information.
\end{enumerate}

\subsubsection{LLM-Based Semantic Text Canonicalization}
\label{LLM}
Textual descriptions associated with narrative scroll paintings are often noisy, stylistically
heterogeneous, and semantically under-specified.
To reduce lexical variance while preserving high-level semantic intent, we employ a Large
Language Model as a deterministic semantic canonicalization operator that rewrites
raw text into a compact, concept-centric representation before multimodal fusion.

We adopt the LLaMA family of autoregressive decoder-only transformers, which model a token
sequence $\mathbf{x} = (x_1,\dots,x_T)$ via causal language modeling:
$P(\mathbf{x})
=
\prod_{t=1}^{T}
P(x_t \mid x_1,\dots,x_{t-1})$,
implemented using multi-head causal self-attention.
LLaMA follows a pre-normalized transformer architecture with RMSNorm, rotary positional
embeddings (RoPE, \cite{Su_24r}), and SwiGLU feed-forward blocks (\cite{Shazeer_20y}), yielding improved stability and semantic
expressivity under constrained inference budgets.
We adopt TinyLlama (\cite{Zhang_24ll}) specifically because its compact parameter count (1.1B) permits inference within the memory constraints of our compute environments without requiring quantization, while its LLaMA-2 architecture retains sufficient semantic expressivity for the canonicalization task, which requires entity recognition and lexical normalization rather than open-ended generation.
Let $\mathbf{h}^{(0)}$ denote the input token embeddings augmented with RoPE.
Each transformer layer $\ell \in \{1,\dots,L\}$ computes
$\mathbf{h}^{(\ell)}
=
\mathrm{Block}^{(\ell)}\!\left(\mathbf{h}^{(\ell-1)}\right)$,
and the final hidden states parameterize a categorical distribution over the vocabulary via
a linear projection and softmax.

In our pipeline, the language model is not used as a direct feature encoder.
Instead, we formalize it as a deterministic rewriting function
$\mathcal{R}_{\mathrm{LLaMA}} :
\mathcal{X}_{\mathrm{text}}
\rightarrow
\widetilde{\mathcal{X}}_{\mathrm{text}}$,
which maps raw panel descriptions to semantically normalized textual forms that emphasize
entities, actions, and symbolic motifs while suppressing surface-level linguistic noise.
For each panel $i$, the rewritten text is given by
$\tilde{x}_i^{\mathrm{text}}
=
\mathcal{R}_{\mathrm{LLaMA}}\!\left(x_i^{\mathrm{text}}\right)$.

The canonicalized text $\tilde{x}_i^{\mathrm{text}}$ is subsequently passed on to downstream
multimodal encoders (SigCLIP and multimodal VAE), where it serves as the linguistic input
for contrastive alignment and latent fusion.
By enforcing semantic consistency at the textual level, this rewriting stage reduces
cross-modal noise and improves the quality of learned image-text representations without
altering the downstream encoder architectures.

\subsubsection{Contrastive Language Image Pretraining (CLIP) models and their variants}
Following semantic canonicalization of textual descriptions, we learn a joint image--text
embedding space via contrastive vision-language pretraining.
The goal is to embed visual panels and their associated texts into a shared latent space
such that semantically aligned image-text pairs are geometrically proximate, while
mismatched pairs are separated. Let
$\mathcal{D}=\{(I_i,T_i)\}_{i=1}^{N}$
denote a corpus of paired images $I_i\in\mathcal{I}$ and texts $T_i\in\mathcal{T}$.
We learn dual encoders
$f_{\theta}:\mathcal{I}\rightarrow\mathbb{R}^{d}$,
\qquad
$g_{\phi}:\mathcal{T}\rightarrow\mathbb{R}^{d}$,
which map each modality into a common $d$-dimensional embedding space.
The architecture follows a dual-tower design: encoders are trained jointly but operate
independently at inference time.

Given a minibatch
$\mathcal{B}=\{(I_i,T_i)\}_{i=1}^{B}$,
we obtain $\ell_2$-normalized embeddings
$\mathbf{x}_i=\frac{f_{\theta}(I_i)}{\|f_{\theta}(I_i)\|_2}$,
\qquad
$\mathbf{y}_i=\frac{g_{\phi}(T_i)}{\|g_{\phi}(T_i)\|_2}$,
\qquad
$\mathbf{x}_i,\mathbf{y}_i\in\mathbb{R}^{d}$.
Pairwise image--text similarity is measured by cosine similarity,
$s_{ij}=\mathbf{x}_i^{\top}\mathbf{y}_j$,
scaled by a temperature parameter $t>0$,
$\ell_{ij}=t\,s_{ij}$.
The CLIP objective treats image-to-text and text-to-image matching as symmetric classification problems.
Define
$p^{I\rightarrow T}_{ij}
=
\frac{\exp(\ell_{ij})}{\sum_{k=1}^{B}\exp(\ell_{ik})}$,
\qquad
$p^{T\rightarrow I}_{ij}
=
\frac{\exp(\ell_{ij})}{\sum_{k=1}^{B}\exp(\ell_{kj})}$.
The resulting contrastive loss is
\begin{equation*}
\mathcal{L}_{\mathrm{CLIP}}
=
-\frac{1}{2B}
\sum_{i=1}^{B}
\left(
\log p^{I\rightarrow T}_{ii}
+
\log p^{T\rightarrow I}_{ii}
\right),
\label{eq:clip_softmax_loss}
\end{equation*}
which maximizes agreement between matched image--text pairs while implicitly using all
other samples in the batch as negatives.

\noindent \textbf{Multimodal Embedding Construction: } Let $\mathcal{M}$ denote a trained CLIP-family encoder with parameters
$(\theta,\phi)$.
For a panel $(I_i,T_i)$, we define a unified multimodal embedding operator
$\mathcal{E}_{\mathcal{M}}:(\mathcal{I},\mathcal{T})\rightarrow\mathbb{R}^{d}$,
given by simple feature fusion of normalized image and text embeddings:
$\mathbf{e}_i^{(\mathcal{M})}
=
\mathcal{E}_{\mathcal{M}}(I_i,T_i)
=
\frac{1}{2}
\left(
\mathbf{x}_i+\mathbf{y}_i
\right)$.
\label{eq:clip_fusion_mean}

Textual descriptions vary substantially in length and narrative structure.
To obtain representations that are robust to localized phrasing effects, we model text
encoding as an expectation over multiple subcontexts.
Given a text sequence $T_i$ of length $|T_i|$, we sample $C$ contiguous substrings
$\{T_i^{(c)}\}_{c=1}^{C}$ of fixed maximum length $s$, drawn uniformly from all valid
substrings.
The final multimodal representation is defined as
\begin{equation*}
\mathbf{e}_i^{(\mathcal{M})}
=
\frac{1}{C}
\sum_{c=1}^{C}
\mathcal{E}_{\mathcal{M}}(I_i,T_i^{(c)}),
\label{eq:clip_chunk_avg}
\end{equation*}
which can be interpreted as a Monte-Carlo approximation to marginalizing the embedding over
textual subcontexts. 

\noindent \textbf{CLIP variants: }We present
three variants of CLIP-style encoders. All of them have an identical architectural backbone and embedding construction
procedure, differing only in the training objective and parameterization used to obtain
the encoder weights.

\noindent \textbf{Pretrained CLIP}
The first variant employs a CLIP model trained on large-scale web image-text pairs using
the softmax-based contrastive objective.
Let $(\theta_0, \phi_0)$ denote the parameters obtained from this pretraining process.
The resulting embeddings
$\mathbf{e}_i^{\mathrm{CLIP\text{-}pre}}
=
\mathcal{E}_{(\theta_0,\phi_0)}(I_i, T_i)$
provide a strong general-purpose multimodal alignment prior and serve as a baseline for
domain-specific adaptation.

\noindent \textbf{Dataset-Specific Softmax-Contrastive Fine-Tuned CLIP}
This variant corresponds to a CLIP model whose parameters
$(\theta_{\mathrm{ft}}, \phi_{\mathrm{ft}})$
are obtained by contrastive training on the image-text pairs drawn exclusively from our dataset.
This data set-specific fine-tuning explicitly aligns the representation space with
the stylistic, iconographic, and linguistic distributions characteristic of narrative
scroll paintings, while retaining the inductive bias of CLIP’s original cross-modal
alignment mechanism.
The resulting multimodal representations,
$\mathbf{e}_i^{\mathrm{CLIP\text{-}ft}}
=
\mathcal{E}_{(\theta_{\mathrm{ft}},\phi_{\mathrm{ft}})}(I_i, T_i)$
exhibit enhanced sensitivity to dataset-specific semantics relative to the pretrained
baseline, thereby providing a stronger foundation for downstream graph-based propagation
and label-conditioned recommendation.

\noindent \textbf{Dataset-Specific Sigmoid-Contrastive Fine-Tuned CLIP (SigCLIP)}
This variant adopts the same dual-encoder CLIP architecture as the preceding models
but replaces the softmax-based contrastive objective with a sigmoid-based pairwise loss
during fine-tuning. Unlike softmax-based contrastive learning, which normalizes similarities across an entire
batch, the sigmoid-contrastive formulation treats each image--text pairing as an
independent binary classification problem.
Let $\mathbf{x}_i$ and $\mathbf{y}_j$ denote the $\ell_2$-normalized image and text
embeddings produced by the encoders.
The similarity logit for a pair $(i,j)$ is defined as
$\ell_{ij}
=
t\,\mathbf{x}_i^{\top}\mathbf{y}_j + b,$
where $t>0$ is a learnable temperature parameter and $b$ is an optional learnable bias
term that stabilizes optimization under large class imbalance.
Pairwise labels are assigned as
\begin{equation*}
z_{ij} =
\begin{cases}
+1, & \text{if } i=j \quad \text{(matched image--text pair)},\\
-1, & \text{if } i\neq j \quad \text{(mismatched pair)}.
\end{cases}
\end{equation*}
The sigmoid-contrastive objective optimized during fine-tuning is then
\begin{equation*}
\mathcal{L}_{\mathrm{SigCLIP}}
=
-\frac{1}{B}
\sum_{i=1}^{B}
\sum_{j=1}^{B}
\log
\sigma\!\left(z_{ij}\,\ell_{ij}\right),
\label{eq:sigclip_loss}
\end{equation*}
where $\sigma(\cdot)$ denotes the logistic sigmoid function.
This loss eliminates batchwise normalization and instead enforces alignment through
independent pairwise decisions, thereby decoupling gradient contributions across samples
within a batch. As the architecture and embedding operator remain unchanged, the resulting multimodal
representations
$\mathbf{e}_i^{\mathrm{SigCLIP\text{-}ft}}
=
\mathcal{E}_{(\theta_{\mathrm{sig}},\phi_{\mathrm{sig}})}(I_i, T_i)$
differ from the pretrained and softmax-fine-tuned variants solely due to the altered
contrastive learning signal.


\subsubsection{Multimodal Variational Autoencoder}
We introduce a multimodal variational
autoencoder (VAE) that learns a probabilistic latent representation jointly grounded
in visual and textual modalities.
Our formulation follows the Auto-Encoding variational Bayes framework (AEVB, \cite{zhihan_22a}), extended to heterogeneous multimodal inputs, and is used as a frozen
feature extractor during downstream evaluation.

For each narrative panel $i$, let $x_i^{\mathrm{img}}$ denote the image modality and
$x_i^{\mathrm{text}}$ the associated textual description.
We posit a continuous latent variable
$\mathbf{z}_i \in \mathbb{R}^{d}$ that generates both modalities.
The joint distribution factorizes as
\begin{equation*}
p(\mathbf{z}_i) = \mathcal{N}(\mathbf{0}, \mathbf{I}),
\qquad
p_\theta(x_i^{\mathrm{img}}, x_i^{\mathrm{text}} \mid \mathbf{z}_i)
=
p_\theta(x_i^{\mathrm{img}} \mid \mathbf{z}_i)\,
p_\theta(x_i^{\mathrm{text}} \mid \mathbf{z}_i),
\end{equation*}
where conditional independence given $\mathbf{z}_i$ enables tractable multimodal modeling
while allowing each decoder to specialize to its respective modality.
Exact posterior inference
$p(\mathbf{z}_i \mid x_i^{\mathrm{img}}, x_i^{\mathrm{text}})$
is intractable.
We therefore introduce modality-specific variational approximations,
each parameterized as a diagonal-covariance Gaussian:
\begin{align*}
q_\phi^{\mathrm{img}}(\mathbf{z}_i \mid x_i^{\mathrm{img}})
&=
\mathcal{N}\!\left(
\boldsymbol{\mu}_i^{\mathrm{img}},
\operatorname{diag}\!\big(\boldsymbol{\sigma}_i^{\mathrm{img}}\big)^2
\right), \\
q_\phi^{\mathrm{text}}(\mathbf{z}_i \mid x_i^{\mathrm{text}})
&=
\mathcal{N}\!\left(
\boldsymbol{\mu}_i^{\mathrm{text}},
\operatorname{diag}\!\big(\boldsymbol{\sigma}_i^{\mathrm{text}}\big)^2
\right).
\end{align*}
These encoders independently extract modality-conditioned evidence about the latent
semantic state $\mathbf{z}_i$.

\subsection{Modality Fusion}
To combine information from different modalities, we adopt a Product-of-Experts (PoE)
formulation.
The fused variational posterior is defined as
\begin{equation*}
q_\phi(\mathbf{z}_i \mid x_i^{\mathrm{img}}, x_i^{\mathrm{text}})
\;\propto\;
q_\phi^{\mathrm{img}}(\mathbf{z}_i \mid x_i^{\mathrm{img}})
\;
q_\phi^{\mathrm{text}}(\mathbf{z}_i \mid x_i^{\mathrm{text}}),
\end{equation*}
which remains Gaussian when all experts are Gaussian.
Let
$\boldsymbol{\Lambda}_i^{(m)}=\operatorname{diag}\!\big(\boldsymbol{\sigma}_i^{(m)}\big)^{-2}$
denote the precision matrix for modality $m\in\{\mathrm{img},\mathrm{text}\}$.
The fused posterior parameters admit the closed-form solution
\begin{align*}
\boldsymbol{\Lambda}_i &= \boldsymbol{\Lambda}_i^{\mathrm{img}} +
\boldsymbol{\Lambda}_i^{\mathrm{text}}, \\
\boldsymbol{\mu}_i &= \boldsymbol{\Lambda}_i^{-1}
\left(
\boldsymbol{\Lambda}_i^{\mathrm{img}} \boldsymbol{\mu}_i^{\mathrm{img}}
+
\boldsymbol{\Lambda}_i^{\mathrm{text}} \boldsymbol{\mu}_i^{\mathrm{text}}
\right), \\
\boldsymbol{\Sigma}_i &= \boldsymbol{\Lambda}_i^{-1}
\end{align*}
This fusion mechanism performs uncertainty-aware integration:
modalities with lower posterior variance exert greater influence on the shared latent
representation. For each panel, we use the posterior mean of the fused distribution as a deterministic
feature representation:
\begin{equation*}
\mathbf{e}_i^{\mathrm{VAE}} = \boldsymbol{\mu}_i
\end{equation*}
This choice eliminates sampling variance while retaining the uncertainty-aware fusion
learned during training.

\subsection{Concept (or Target Label) Learning} Twenty annotators, previously trained, were asked to identify the presence or absence of three different concepts in a panel: (a) animals, (b) trees, and (c) mythological characters. The goal is to find out how good humans are at identifying these concepts. Krippendorff's alpha(\cite{Kripp_04a}) 
is used to measure inter-annotator agreement. The following were the values of alpha for different concepts: (a) Myth. char: 0.69 (b) Tree: 0.85 (c) Animals: 0.69.

\subsection{Graph Structure Learning (GSL)}
\label{GSLearn}
\subsubsection{Learning homogeneous graphs}
To discover latent structures of graphs from panels, we first use the multimodal features to learn $|\mathcal{P} \times \mathcal{P}|$ similarity matrices. Using the hypothesis that similar panels are likely to be of interest more often to the end-user than dissimilar ones \cite{McPherson_03a}, the semantic relationship between panels is quantified by using similarity measures such as cosine similarity (\cite{Wang_20a}, \cite{Zhu_21}, \cite{Zhang_20a}, \cite{Chen_20a}) kernel-based functions (\cite{Li_2018}), or attention mechanisms (\cite{Chen_20a}). In this work, the parameter-free cosine similarity is used so that the similarity $S_m = \frac{(e_i^m)^T e_j^m}{\parallel e_i^m \parallel \parallel e_j^m \parallel}$, where $e_i^m, e_j^m$ represents multimodal feature vectors. Since the adjacency matrix of a graph has positive values and a fully connected complete graph is often undesirable, we suppress the negative values in $S_m$. In addition, motivated by the fact that real-world graphs are noisy and can contain many task-unrelated edges, we apply a sparsity constraint. Borrowing an idea from $\epsilon$-proximity thresholding graphs ($\epsilon$-graphs \cite{Bentley_77}), we create an edge between two nodes if their similarity is smaller than a predefined, data-driven, threshold $\epsilon$.

\subsubsection{Transductive versus Inductive Graph Learning}

Training neural networks on graphs can be done in either the transductive or inductive setting (\cite{Xu_20a,Lachaud_23a}). In the transductive setting, the model has access to the test features in the training phase. In the inductive setting, the test data remain unseen. We explore the differences in inductive versus transductive learning primarily because field studies conducted at different times from disparate geographic locations have added data to our corpus at a steady pace, necessitating re-training of models and relying on inductive learning.

\subsection{Node Classification } 
\label{GCN}

Let $\mathcal{G} = (A, X)$ denote a graph where $A \in R^{N \times N}$ is the adjacency matrix and $X \in R^{N \times F}$ is the node feature matrix with the $i$-th entry $x_i \in R^F$ denoting the feature of the node $v_i$. Given $\mathcal{G}$, the goal of GSL is to simultaneously learn an adjacency matrix $A^{*}$ and its corresponding node representations 
for downstream tasks. A GSL model has two trainable components: (1) a GNN encoder $f_{\theta}$ that takes a graph as input and generates an embedding. (2) a structure learning module $g_{\phi}$ that contains the edge connectivity information of the graph. 

The model parameters $(\theta, \phi)$ are trained with the following learning objective:
\begin{equation}
\label{eqnGNNa}
\mathcal{L} = \mathcal{L}_{tsk} (Z^{*}, Y) + \lambda \mathcal{L}_{reg} (A^{*}, Z^{*},\mathcal{G})
\end{equation}
where the first term on the right-hand side represents the original task, the second -- regularizer on the learned graph, and $\lambda$ is a hyperparameter to be tuned. 

Thus, node classification can be used to assign labels (such as the presence/absence of a concept) to nodes in a graph based on the properties of nodes and the relationships between them.  The labels assigned to nodes (panels) are used for relevance estimation in the recommendation system. 

The formulation used in Equation~\ref{eqnGNNa} relies on the assumption that connected edges in a graph are likely to share the same label. This assumption may restrict modeling capacity as graph edges need not encode similarity but may contain additional information. To address this, Graph Auto Encoders (GAEs \cite{Kipf_17a}) have been proposed, wherein the graph structure is modeled using a neural network $f(X,A)$. We call this GeMi GAE in our design and empirical analysis.

Marrying ideas from deep neural networks and approximate Bayesian inference, it is possible to further enhance the framework for encoding graph structures, using the Variational Graph Auto Encoder (VGAE, \cite{kipf_16a}).
VGAEs are capable of learning interpretable latent representations for undirected graphs. This can be demonstrated by using a Graph Convolutional Network (GCN) encoder and an inner product decoder. Assume stochastic latent variable $z_i$ summarized in an $N \times D$ matrix $Z$. The inference model parameterized by a two-layer GCN is given by
\begin{equation*}
q(z|X,A) = \prod_{i=1}^{N} q(z_i|X, A)
\end{equation*}
with $q(z_i|X,A)=\mathcal{N}(z_i|\mu_i, diag(\sigma_i^2))$
Also, $\mu = GCN_{\mu}(X,A)$ is the matrix of mean vectors of $\mu_i$ and $log \sigma = GCN_{\sigma} (X,A)$. The two-layer GCN is defined as follows:
\begin{equation*}
GCN (X,A) = \tilde{A} \quad ReLU (\tilde{A} X W_0) W_1
\end{equation*} with $W_i$ being the weight matrices; $GCN_{\mu}$ and $GCN_{\sigma}$ share the first layer of parameters $W_0$ and $\tilde{A}$ is a symmetrically normalized adjacency matrix. The generative model is given by:
\begin{equation*}
p(A|Z) = \prod_{i=1}^{N} \prod_{j=1}^{N} p(A_{ij}|z_i,z_j)
\end{equation*}
with $p(A_{ij}=1|z_iz_j)=\sigma(z_i^Tz_j)$, $\sigma(.)$ is the logistic sigmoid function. Finally, the learning step involves optimizing the variational lower bound $\mathcal{L}$ w.r.t the variational parameters $W_i$:
\begin{equation*}
\mathcal{L}= E_{q(Z|X,A)}[log p (A | Z )] - KL[q(Z|X,A) || p(Z)]
\end{equation*}
where $KL[q||p]$ is the Kullback-Leibler divergence between $q(.)$ and $p(.)$. This model is referred to as GeMi VGAE in our architecture.





The three graph encoder variants in GeMi — GCN, GAE, and VGAE — are chosen to span a principled spectrum of inductive biases. GCN provides a discriminative baseline that propagates supervised label signals directly through the graph. GAE introduces an unsupervised reconstruction objective, allowing the model to leverage unlabeled structural information without relying solely on sparse concept labels. VGAE extends this with a probabilistic latent space, which is particularly well-suited to our setting: the scroll painting dataset is small and noisy, and a variational formulation explicitly models uncertainty in node representations. Together, these three architectures allow us to study the trade-off between supervised signal strength, structural self-supervision, and latent uncertainty modeling — questions that are non-trivial to answer a priori for a novel dataset of this kind.

\subsection{Strategies for Handling Class Imbalance}
\label{subsec:balancing}

The concepts to be learned through node classification may be imbalanced (for example, \textit{Tree}). In a graph-based recommendation system, this skew may cause (i) gradients to be dominated by majority labels, (ii) decision thresholds to shift toward predicting negatives for rare labels, and (iii) graph neighborhoods to reflect dense majority clusters rather than minority structure (\cite{Markwald_2024a, Zhao_21f}). To counter these effects, we employ balancing strategies at the following three levels: 
\begin{itemize}
    \item graph structure
    \item loss reweighting and hard-example emphasis
    \item optimization regularization/stabilization
\end{itemize}
Details about these methods are presented in Section~\ref{EmpResults}.

\subsection{Incorporating Collaborative Filtering (CF) Methods}
GeMi separates the process of multimodal feature extraction from user-panel interaction information in recommendation tasks. After learning panel representations, the user-panel interactions are incorporated with downstream collaborative filtering tasks. This allows collaborative filtering to be served as a plug-and-play module in the overall architecture.

\section{Empirical Results}
\label{EmpResults}

\subsection{Aims}
\label{aims}
Our aim is to answer the following research questions:
\begin{itemize}

\item \textbf{RQ1: (Baseline Comparison)} 
How does our system (GeMi) perform when compared to the state-of-the-art graph-based recommendation systems built using homogeneous item-item graphs and other heterogeneous user-item graphs used in collaborative filtering settings?
\item \textbf{RQ2: (Feature Generation)} Do LLM-enhanced vision-language features contribute to better model performance of the recommendation system?
\item \textbf{RQ3: (Graph Structure Learning - Inductive versus Transductive training)} Does an inductive or transductive training procedure provide better results for recommendation?

\end{itemize}
\subsection{Materials}

\begin{figure}
     \centering
     \begin{subfigure}[b]{0.45\textwidth}
         \centering
         \includegraphics[width=\textwidth]{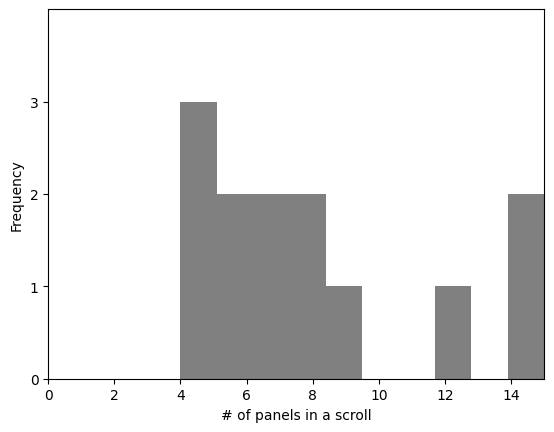}
         \caption{Phase 1}
         \label{fig:ph1}
     \end{subfigure}
     \hfill
     \begin{subfigure}[b]{0.45\textwidth}
         \centering
         \includegraphics[width=\textwidth]{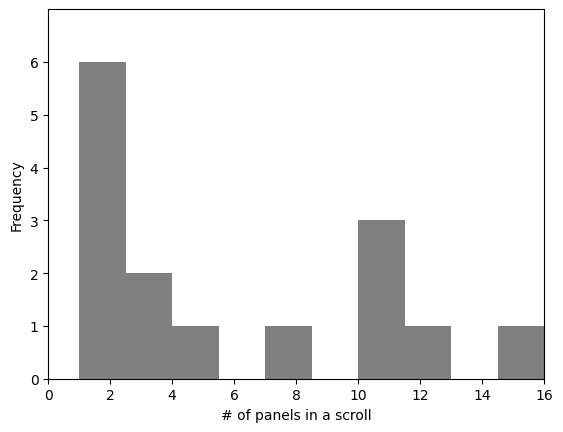}
         \caption{Phase 2}
         \label{fig:ph2}
     \end{subfigure}
        \caption{Distribution of the number of panels per scroll in the two phases of data collection from field surveys.}
        \label{fig:dataCol}
\end{figure}

 \subsubsection{Data}
 We conducted a series of experiments on a meticulously curated dataset of panels and associated songs collected during two phases of field studies. There are a total of 106 panels from 13 scrolls, paired with 13 distinct songs collected in Phase 1. In Phase 2, 15 distinct scrolls with a total of 83 panels were collected. Figures~\ref{fig:ph1} and Figure~\ref{fig:ph2} show the distribution of the number of panels per scroll collected in these two phases. Although the data from phase 1 are well aligned (every panel illustrates an excerpt of a song), only $26.67\%$ of the scrolls in phase 2 have associated songs. 
Each panel in the data set is annotated with the presence or absence of three key concepts(labels = $0$ or $1$): tree, animal and mythology. 

 \subsubsection{Machines}

 All inductive experiments were conducted on a university-managed High Performance Computing (HPC) compute node under a single-GPU configuration. Unless otherwise specified, each run reserved 8 CPU cores, 64 GB RAM, and 1 GPU per job. The system supported a software stack with Jupyter, PyTorch 3, and CUDA, and computations were run on AVX-512 capable CPUs.
All transductive experiments were conducted using Google Colab as the execution environment and were run with Python~3 and a single NVIDIA Tesla T4 GPU as the hardware accelerator, providing 16 GB of GPU memory. 

  \subsubsection{Baseline Algorithms}
  \label{baseAlgos}

  We compare the performance of GeMi to the state-of-the-art transductive and inductive graph-based recommendation systems. These algorithms are briefly described in this section.
  
   \begin{itemize}
    \item Transductive homogenous baselines
        \begin{enumerate}
            \item LATTICE (\cite{lattice_21}) constructs KNN item graphs for each modality and refines them into a skip-connected latent adjacency that is softmax-fused across modalities. A LightGCN-style propagation is then applied on the fused graph to strengthen BPR-optimized CF backbones.
            \item PMGT (\cite{pmgt_21}) performs pretraining on an item multimodal graph through a Transformer encoder, optimized with graph structure reconstruction and masked node feature reconstruction.
            \item HUIGN (\cite{huign_22}) models multi-level user intents through hierarchical graph convolution learned from a co-interacted item graph, forming a fine-to-coarse hierarchy of latent intents that jointly refine user and item embeddings.
            \item DualGNN (\cite{dualgnn_23}) applies simplified GNN propagation on modality-specific user–item bipartite graphs to learn per-modality user preferences, which are then attention-fused and propagated over a user co-occurrence graph to handle modality missing in user representations.
            \item FREEDOM (\cite{freedom_23}) improves multimodal recommendation by freezing a KNN-based item–item graph constructed from raw multimodal features while denoising the user–item graph through degree-sensitive edge pruning to suppress noisy interaction.
            \item MICRO (\cite{micro_23}) learns modality-specific item–item graphs from multimodal features. It refines embeddings with graph convolutions and aligns modalities through a contrastive fusion objective. 
            \item DGVAE (\cite{dgvae_24}) uses a disentangled graph variational auto-encoder to learn interpretable multimodal user preferences, by encoding both user–item interactions and user–word preferences on a fixed item–item graph.
        \end{enumerate}
\noindent Since all of the above algorithms construct homogeneous item-item graphs, they are closest in spirit to the GeMi system.

    \item Transductive heterogeneous baselines
        \begin{enumerate}
            \item MambaRec (\cite{mambarec})  integrates local alignment through a Dilated Refinement Attention Module (DREAM) with multi-scale convolutions and dual attention, and global modality regularization through Maximum Mean Discrepancy (MMD) and contrastive objectives to enforce cross-modal semantic consistency.
            \item PGL (\cite{pgl}) extracts a principal user–item subgraph (via global-aware truncation or local-aware sampling) to preserve local structural information, with InfoNCE-based feature masking, and also leveraging multimodal item–item graphs.
            \item HPMRec (\cite{hpmrec}) encodes user–item representations with multi-component hypercomplex embeddings. It performs nonlinear cross-modal fusion through hypercomplex multiplication and uses prompt-aware compensation to dynamically align components.
            \item COHESION (\cite{cohesion}) is a composite GCN that integrates graphs user-item, user–user, and item-item, refines embeddings via behavior modality, and then carefully fuses representations of the modality, with an adaptive BPR objective to balance learning.
            \item SMORE (\cite{smore}) fuses visual and textual features in the frequency domain, with adaptive spectral filtering to suppress modality-specific noise before fusion. It models both uni-modal and fused preferences through multi-view graph learning and a modality-aware preference module.
            \item CMDL  (\cite{CMDL}) disentangles multimodal representations into modality-invariant and modality-specific components using a novel mutual-information–based disentanglement regularization.
        \end{enumerate}
   
    \item Inductive baselines: 
    All the inductive baselines are studied both in the homogeneous (item-item) and heterogeneous (user-item) settings. 
        \begin{enumerate}
            \item GraphSAGE (\cite{Hamilton_18a}): In the homogeneous setting, this algorithm first constructs an item-item graph using cosine similarity on item embeddings and a fixed percentile threshold; it then trains a two-layer GraphSAGE encoder to learn the aggregation function over node features. In the heterogeneous setting, it constructs a user-item bipartite graph through neighborhood aggregation.
            \item PinSAGE\footnote{Our implementation of the PinSAGE algorithm is available from \url{https://github.com/Haimonti/Folk-Art-Recommendation-System/tree/Inductive_Jin/Inductive_settings} and is an improved version of \cite{Cao_25a}.}(\cite{Cao_25a, Ying_18p})
            It replaces purely feature-based neighborhoods for node aggregation with random-walk–based neighborhoods by precomputing co-visitation counts and rewiring the item-item graph with top-k most frequently visited neighbors. 
            \item GATNE-I (\cite{Cen_2019}) learns item embeddings on a single-modality KNN graph using a skip-gram objective with negative sampling. 
        \end{enumerate}

    \end{itemize}

 \subsubsection{User-Preference Matrix Generation}
  \label{user-pref}
  
We simulate user preferences by constructing a synthetic user-preference matrix derived from the training corpus through Monte-Carlo sampling. Let the training set consist of $N_{\mathrm{tr}}$ panels.
Each panel $i$ is annotated with a binary multi-label indicator vector $\mathbf{y}_i$, where
$\mathbf{y}_i
=
\bigl( y_{i1}, \dots, y_{i|\mathcal{L}|} \bigr)
\in \{0,1\}^{|\mathcal{L}|}$,
over a fixed semantic label set
$\mathcal{L}$, where $|\mathcal{L}| = 3$ corresponds to the categories
\emph{Animal}, \emph{Mythology}, and \emph{Tree}.
  
\noindent \textbf{Synthetic User Construction: }We generate a population of $U$ synthetic users, each intended to represent a plausible
latent preference profile over the label space.
For each synthetic user $u \in \{1,\dots,U\}$, we sample a subset of $K$ distinct training
items,
$\mathcal{I}_u \subset \{1,\dots,N_{\mathrm{tr}}\},
\qquad |\mathcal{I}_u| = K, \qquad 1 \le K \le N_{\mathrm{tr}}$
The interaction sets are sampled independently according to the uniform distribution
over all $K$-subsets of the training index set:
\begin{equation*}
\mathcal{I}_u
\;\sim\;
\mathrm{Unif}
\!\left(
\left\{
\mathcal{S} \subseteq \{1,\dots,N_{\mathrm{tr}}\}
\;\middle|\;
|\mathcal{S}| = K
\right\}
\right)
\end{equation*}
This sampling scheme treats each training item as an equally likely candidate interaction
and enforces diversity within a user profile by prohibiting repeated items.
When the training set size is smaller than $K$, the effective interaction count is
implicitly capped at $N_{\mathrm{tr}}$ to ensure feasibility.

\noindent \textbf{Preference Estimation: }Given the sampled interaction set $\mathcal{I}_u$, we estimate the latent interests of
user $u$ by aggregating the semantic labels associated with the selected items.
Specifically, we compute the empirical label frequency vector
$\hat{\mathbf{f}}_u
=
\frac{1}{K}
\sum_{i \in \mathcal{I}_u}
\mathbf{y}_i
\;\in\;
[0,1]^{|\mathcal{L}|}$,
where each component $\hat{f}_{u\ell}$ represents the fraction of sampled interactions
that exhibit label $\ell$. To obtain a discrete and interpretable preference representation, we apply a fixed
threshold $\tau \in (0,1)$ to the empirical frequencies, yielding a binary preference
vector
$\mathbf{p}_u
=
\mathbb{I}\!\left[ \hat{\mathbf{f}}_u \ge \tau \right]
\;\in\;
\{0,1\}^{|\mathcal{L}|}$,
where the indicator function is applied elementwise.
Equivalently, a label is deemed preferred by user $u$ if it appears in at least a
$\tau$-fraction of the sampled interactions.
Equivalently, for each label
$\ell \in \{1,\dots,|\mathcal{L}|\}$,
\begin{equation*}
p_{u\ell}
=
\begin{cases}
1,
& \text{if }
\frac{1}{K}
\sum_{i \in \mathcal{I}_u}
y_{i\ell}
\ge \tau, \\[6pt]
0,
& \text{otherwise}
\end{cases}
\end{equation*}
This formulation naturally adapts to the interaction budget $K$, ensuring that preference
activation reflects repeated evidence rather than isolated occurrences.
\noindent \textbf{User-Preference Matrix: } Aggregating the preference vectors across all synthetic users yields the user-preference
matrix
$\mathbf{P}
=
\begin{bmatrix}
\mathbf{p}_1^{\top} \\
\mathbf{p}_2^{\top} \\
\vdots \\
\mathbf{p}_U^{\top}
\end{bmatrix}
\;\in\;
\{0,1\}^{U \times |\mathcal{L}|}$.
Each row of $\mathbf{P}$ encodes a distinct synthetic user’s inferred interests over the
semantic label space, while each column corresponds to a specific label category.
The total number of training item appearances across all profiles is
$UK$, and assuming approximately uniform sampling, each training item participates in
roughly $\frac{UK}{N_{\text{tr}}}$ synthetic profiles on average, promoting broad
coverage of the training distribution.

\subsection{Methods}
\FloatBarrier
The following details are relevant:

\begin{itemize}
\item \textbf{Parameters of GeMi variants: } Table~\ref{tr_ablation} summarizes the architectural and optimization configurations of the three GeMi variants, namely GeMi-GCN, GeMi-GAE, and GeMi-VGAE, under the transductive protocol. In all cases, graph construction is based on cosine-similarity k-nearest-neighbor connectivity, supervised learning employs focal binary cross-entropy with logits, optimization is performed using Adam, and regularization includes dropout, edge dropout, and gradient clipping. Evaluation is conducted using fixed synthetic user profiles with $k=5$ interactions and a preference threshold of 0.2.

To ensure comparability across variants, several core hyperparameters are intentionally aligned. These include dropout (0.20), optimizer (Adam), gradient clipping with maximum norm 2.0, kNN-based graph construction (with variant-specific $k$ values of 30, 30, and 25), edge dropout (0.10), ReLU activation in graph and head layers, and focal loss parameters $\gamma=2.0$ and $\alpha=0.25$. Furthermore, the same recommendation evaluation setup ($k=5$ interactions and preference threshold as 0.2) is maintained across all models.

\begin{table*}[!h]
\centering
\small
\setlength{\tabcolsep}{10pt}
\begin{tabular}{lp{0.78\textwidth}}
\toprule
\multicolumn{1}{c}{\textbf{GeMi Variant}} & \multicolumn{1}{c}{\textbf{Parameters}} \\
\midrule
GeMi-GCN &
Graph encoder: 2-layer Graph Convolutional Network;
GCN hidden dimension: 128;
GCN dropout: 0.20;
Training epochs: 450;
Optimizer: Adam;
Activation: ReLU;
Learning rate: $3\times10^{-4}$;
Weight decay: $2\times10^{-3}$;
Gradient clipping: max norm 2.0;
Loss (balanced): focal binary cross-entropy with logits;
Focal parameters: $\gamma=2.0$, $\alpha=0.25$;
kNN neighbors ($k$): 30;
Tree-edge augmentation neighbors ($k_{tree}$): 25;
Tree augmentation max nodes: 2500;
Edge dropout: 0.10;
$k=5$ interactions;
Preference threshold: 0.2
\\
GeMi-GAE &
Graph encoder: Graph Auto-Encoder (GAE) with a 2-layer GCN encoder;
Hidden dimension: 128;
Latent dimension ($d_z$): 64;
Dropout: 0.20;
Training epochs: 400;
Optimizer: Adam;
Activation: ReLU;
Learning rate: $2\times10^{-3}$;
Weight decay: $2\times10^{-4}$;
Gradient clipping: max norm 2.0;
kNN neighbors ($k$): 30;
Edge dropout: 0.10;
Self-supervised loss: GAE reconstruction loss $\mathcal{L}_{rec}$;
Supervised loss: focal binary cross-entropy with logits;
Focal parameters: $\gamma=2.0$, $\alpha=0.25$;
Joint objective: $\mathcal{L}=\mathcal{L}_{rec} + \lambda_{sup}\mathcal{L}_{sup}$ with $\lambda_{sup}=0.60$;
Augmentation neighbors: $k_{animal}=18$, $k_{myth}=18$, $k_{tree}=35$;
Augmentation max nodes per label: 2500;
$k=5$ interactions;
Preference threshold: 0.2
\\
GeMi-VGAE &
Graph encoder: Variational Graph Auto-Encoder (VGAE) with GCN-based parameterization;
Hidden dimension: 256;
Latent dimension ($d_z$): 128;
Dropout: 0.20;
Log-standard-deviation clamp: $\log\sigma \in [-10,\,10]$;
Training epochs: 500;
Optimizer: Adam;
Activation: ReLU;
Learning rate: $3\times10^{-3}$;
Weight decay: $5\times10^{-4}$;
Gradient clipping: max norm 2.0;
kNN neighbors ($k$): 25;
Tree augmentation neighbors: $k_{tree}=25$;
Tree augmentation max nodes: 2500;
Edge dropout: 0.10;
Reconstruction loss: VGAE reconstruction loss $\mathcal{L}_{rec}$;
KL regularization: VGAE KL loss $\mathcal{L}_{KL}$;
KL annealing: $\beta_{KL}(t)$ linearly increased to $\beta_{\max}=1.0$;
Self-supervised VGAE objective: $\mathcal{L}_{vgae}=\mathcal{L}_{rec}+\beta_{KL}\mathcal{L}_{KL}$;
Supervised loss: focal binary cross-entropy with logits;
Focal parameters: $\gamma=2.0$, $\alpha=0.25$;
Joint SSL objective: $\mathcal{L}=\mathcal{L}_{vgae}+\lambda_{ssl}\mathcal{L}_{sup}$ with $\lambda_{ssl}=0.60$;
$k=5$ interactions;
Preference threshold: 0.2
\\
\bottomrule
\end{tabular}
\caption{Parameters of GeMi Variants}
\label{tr_ablation}
\end{table*}

\item \textbf{Balanced Graph Structure Learning using Symmetric KNN: }Instead of constructing edges using a global similarity threshold (which tends to over-connect dense majority regions and under-connect rare regions), we construct a \emph{symmetric k-nearest-neighbor (kNN) graph} over all nodes. Let $\mathbf{X}\in\mathbb{R}^{N\times d}$ be node features. We first $\ell_2$-normalize:
$\tilde{\mathbf{x}}_i = \frac{\mathbf{x}_i}{\|\mathbf{x}_i\|_2+\varepsilon}$.
We compute cosine similarities $s_{ij}=\cos(\tilde{\mathbf{x}}_i,\tilde{\mathbf{x}}_j)$ and, for each node $i$, select its top-$k$ neighbors:
$\mathcal{N}_k(i)=\operatorname*{arg\,topk}_{j\neq i}\; s_{ij}$.
We then add \emph{bidirectional} edges (symmetric closure):
$A_{ij}= \mathbb{I}[j\in\mathcal{N}_k(i)] \;\lor\; \mathbb{I}[i\in\mathcal{N}_k(j)]$.
Optionally, we apply a similarity floor $s_{ij}\leftarrow \max(s_{ij}, 0)$ to avoid negative edges. Symmetric kNN enforces roughly uniform local neighborhood size, improving representation learning for sparse concepts by ensuring every node participates in comparable amounts of graph propagation.

  \item \textbf{Class Imbalance Handling Strategies: }We employ several different strategies for handling class imbalance. These include:
    \begin{enumerate}
    
        \item Minority-Aware Edge Augmentation for the \textit{Tree} Label: To explicitly address extreme sparsity in the \textit{Tree} class, we add extra edges among \textit{Tree-positive} training nodes. Let $\mathcal{V}_{\text{tree}}=\{i \mid y_{i,\text{tree}}=1\}$ be indices of nodes labeled positive for Tree (restricted to train nodes in practice). We compute cosine similarities within $\mathcal{V}_{\text{tree}}$ and connect each such node to its top-$k_{\text{tree}}$ neighbors:
$\mathcal{N}^{\text{tree}}_{k_{\text{tree}}}(i)=
\operatorname*{arg\,topk}_{j\in \mathcal{V}_{\text{tree}},\ j\neq i}\; \cos(\tilde{\mathbf{x}}_i,\tilde{\mathbf{x}}_j)$.
We then union these edges with the base kNN graph:
$\mathcal{E} \leftarrow \mathcal{E}\ \cup\ \mathcal{E}_{\text{tree}}$.
For computational feasibility and to prevent over-connection, we cap the number of Tree nodes used in augmentation (random subsampling if needed). Adding intra-minority edges increases within-class information flow, stabilizes embeddings for minority positives, and reduces the chance that these nodes are overwhelmed by messages dominated by majority neighborhoods.

        \item Edge Dropout to Reduce Majority-Connectivity Overfitting:
During each training epoch, we apply \emph{edge dropout} by randomly removing edges with probability $p_e$. For an edge set $\mathcal{E}$, we sample Bernoulli masks $\delta_e \sim \text{Bernoulli}(1-p_e)$ and keep:
$\mathcal{E}'=\{e\in\mathcal{E}\mid \delta_e=1\}$.
Edge dropout acts as structural regularization, encouraging robustness to missing edges and preventing reliance on a small subset of dominant connections, which benefits minority generalization.

    \item Class-Balanced Reweighting via Positive Weights: We compute per-label positive weights from the \emph{training} labels to counteract skewed prevalence. For label $\ell$, let $P_\ell=\sum_{i\in \text{train}} y_{i\ell}$ be positives and $N_\ell = n_{\text{train}}-P_\ell$ negatives. We set:
$w_\ell = \frac{N_\ell}{P_\ell}$,
(with clamping to avoid division by zero), and use weighted BCE-with-logits:
\begin{equation*}
\mathcal{L}_{\text{WBCE}}
=
\frac{1}{n_{\text{train}}}
\sum_{i\in \text{train}}
\sum_{\ell}
\Big(
- w_\ell\, y_{i\ell}\log \sigma(z_{i\ell})
- (1-y_{i\ell})\log(1-\sigma(z_{i\ell}))
\Big),
\end{equation*}
where $z_{i\ell}$ are logits and $\sigma(\cdot)$ is the sigmoid. 
Without reweighting, the empirical risk is dominated by abundant negatives and frequent labels, causing the optimizer to favor conservative predictions (especially for rare labels). $w_\ell$ increases the gradient contribution of minority positives, effectively equalizing the cost of misclassifying rare positives versus frequent ones.

    \item Focal Loss to Emphasize Hard and Minority Examples: 
We further incorporate \emph{focal modulation} on top of BCE-with-logits. Let $p_{i\ell}=\sigma(z_{i\ell})$. Define
\begin{equation*}
p_{t} =
\begin{cases}
p_{i\ell}, & y_{i\ell}=1,\\
1-p_{i\ell}, & y_{i\ell}=0.
\end{cases}
\end{equation*}
We apply the focusing factor $(1-p_t)^\gamma$ and an $\alpha$-balance term:
$\alpha_t = \alpha\, y_{i\ell} + (1-\alpha)(1-y_{i\ell})$,
yielding the focal-BCE objective:
\begin{equation*}
\mathcal{L}_{\text{focal}}
=
\frac{1}{n_{\text{train}}}
\sum_{i\in \text{train}}
\sum_{\ell}
\alpha_t\,(1-p_t)^\gamma\ \mathrm{BCEWithLogits}(z_{i\ell},y_{i\ell}; w_\ell).
\end{equation*}
Focal loss reduces weights on well-classified examples and concentrates learning on hard cases—often boundary samples and minority positives—improving minority recall without excessively inflating false positives.

    \item Training Stabilization Under Reweighted Objectives:  Reweighting (pos-weights) and focal modulation can increase gradient variance. We therefore adopt stabilization measures:
    \begin{enumerate}
    \item Gradient clipping: after backpropagation, we clip the global norm
    \begin{equation}
    \|\nabla_\theta \mathcal{L}\|_2 \leftarrow \min(\|\nabla_\theta \mathcal{L}\|_2,\ \tau),
    \end{equation}
    preventing rare-label spikes from destabilizing updates.
    \item Regularization and capacity tuning: we use a larger hidden dimension and dropout in the GCN and employ weight decay, improving generalization under class imbalance.
    \item Edge dropout (above) as a structural regularizer complements loss balancing by preventing topology overfitting.
\end{enumerate}

Balancing strategies intentionally amplify minority gradients; clipping and regularization ensure these amplified gradients do not lead to exploding updates, oscillations, or overfitting, especially in multi-label regimes where some samples contribute to multiple losses.
    \end{enumerate}

 \item \textbf{Transductive and Inductive Graph Learning: }The GeMi system is studied in both the transductive and inductive settings -- for each setting, state-of-the-art baseline algorithms for homogeneous graphs are used for comparison. In addition, in the transductive setting, we also study collaborative filtering style algorithms built on heterogeneous graphs. While the baseline algorithms have been discussed in Section~\ref{baseAlgos}, Table~\ref{tab:TransHomoPara} and Table~\ref{tab:TransHeteroPara} elucidate the parameters used to run them. 

 \item \textbf{Relevance Estimation and Recommendation: }We formalize relevance estimation as a label-conditioned semantic consistency problem between inferred user preferences and recommended items.
Let $\mathcal{D}_{\mathrm{tr}}$ and $\mathcal{D}_{\mathrm{te}}$ denote the training and test
splits, respectively, with cardinalities
$|\mathcal{D}_{\mathrm{tr}}| = N_{\mathrm{tr}}$ and
$|\mathcal{D}_{\mathrm{te}}| = N_{\mathrm{te}}$.
Each item (panel) $i \in \mathcal{D}_{\mathrm{tr}} \cup \mathcal{D}_{\mathrm{te}}$
is associated with a binary multi-label annotation
\begin{equation}
\mathbf{y}_i = (y_{i1}, \dots, y_{i|\mathcal{L}|}) \in \{0,1\}^{|\mathcal{L}|},
\end{equation}
where $\mathcal{L}$ denotes the fixed semantic label set
(with $|\mathcal{L}| = 3$ in our setting).
The label indicators are treated as ground-truth semantic descriptors of item content. We consider a family of embedding models indexed by $m \in \mathcal{M}$.
For each model $m$, items in the training and test splits are embedded into a
$d_m$-dimensional latent space,
$\mathbf{e}^{(m)}_i \in \mathbb{R}^{d_m},
\qquad
i \in \mathcal{D}_{\mathrm{tr}} \cup \mathcal{D}_{\mathrm{te}}$
\begin{enumerate}
        \item Synthetic User Preference Model: Let $u \in \{1,\dots,U\}$ denote a user.
    Each user is associated with an inferred binary preference vector
$\mathbf{p}_u = (p_{u1}, \dots, p_{u|\mathcal{L}|}) \in \{0,1\}^{|\mathcal{L}|}$,
where $p_{u\ell} = 1$ indicates that user $u$ is inferred to prefer semantic label
$\ell \in \mathcal{L}$.
The construction of $\mathbf{p}_u$ follows the Monte Carlo sampling and thresholding
procedure described in Section~\ref{user-pref}, and is treated as fixed for the
remainder of the evaluation.

        \item User Representation in Latent Space: Given a synthetic user $u$, we associate a latent user representation
$\mathbf{e}^{(m)}_u \in \mathbb{R}^{d_m}$ by aggregating the embeddings of the items sampled to generate the user profile.
This representation is interpreted as a point estimate of the user's latent semantic interest under model $m$.

        \item Scoring and Ranking Function:  For a given model $m$ and synthetic user $u$, each test item
$t \in \mathcal{D}_{\mathrm{te}}$ is assigned a relevance score
$s^{(m)}(u,t)
=
\mathrm{sim}\!\left(
\mathbf{e}^{(m)}_u,\,
\mathbf{e}^{(m)}_t
\right),$
where $\mathrm{sim}(\cdot,\cdot)$ denotes cosine similarity,
$\mathrm{sim}(\mathbf{x},\mathbf{z})
=
\frac{\mathbf{x}^\top \mathbf{z}}
{\|\mathbf{x}\|_2\,\|\mathbf{z}\|_2}$
The recommended list of size $K_{\mathrm{rec}}$ is defined as
$\mathcal{R}^{(m)}_u
=
\arg\max_{\substack{\mathcal{S} \subset \mathcal{D}_{\mathrm{te}} \\ |\mathcal{S}| = K_{\mathrm{rec}}}}
\sum_{t \in \mathcal{S}} s^{(m)}(u,t)$,
i.e., the set of test items attaining the top-$K_{\mathrm{rec}}$ similarity scores.

        \item Label-Conditioned Relevance Definition
Relevance is evaluated independently for each semantic label.
For user $u$, item $t$, and label $\ell$, we define the label-wise relevance indicator
\begin{equation*}
r_{u,t,\ell}
=
\mathbb{I}
\!\left[
p_{u\ell} = 1 \;\wedge\; y_{t\ell} = 1
\right]
\end{equation*}
Thus, a recommended item is deemed relevant for label $\ell$ if and only if
(i) the user is inferred to prefer label $\ell$, and
(ii) the item expresses label $\ell$ in its ground-truth annotation.
Equivalently, for each user $u$ and label $\ell$, we define the label-conditioned
candidate set
\begin{equation*}
\mathcal{C}_{u,\ell}
=
\{ t \in \mathcal{D}_{\mathrm{te}} \mid p_{u\ell} = 1 \;\wedge\; y_{t\ell} = 1 \},
\end{equation*}
and the set of relevant recommendations as
\begin{equation*}
\mathcal{RR}^{(m)}_{u,\ell}
=
\mathcal{R}^{(m)}_u \cap \mathcal{C}_{u,\ell}
\end{equation*}

    \item Evaluation Metric (Precision@K): For each synthetic user $u$, label $\ell$, and model $m$, we compute
label-specific Precision@$K_{\mathrm{rec}}$ as
\begin{equation}
\mathrm{P@}K_{\mathrm{rec}}^{(m)}(u,\ell)
=
\frac{|\mathcal{RR}^{(m)}_{u,\ell}|}{K_{\mathrm{rec}}}
=
\frac{1}{K_{\mathrm{rec}}}
\sum_{t \in \mathcal{R}^{(m)}_u}
r_{u,t,\ell}
\end{equation}
This yields, for each user, a vector of label-wise precision scores
\begin{equation}
\boldsymbol{\pi}^{(m)}_u
=
\big(
\mathrm{P@}K_{\mathrm{rec}}^{(m)}(u,1),
\dots,
\mathrm{P@}K_{\mathrm{rec}}^{(m)}(u,|\mathcal{L}|)
\big)
\in \mathbb{R}^{|\mathcal{L}|}
\end{equation}

Aggregating over the population of synthetic users, we report the empirical mean and
standard deviation
\begin{equation}
\mu^{(m)}_\ell
=
\frac{1}{U}
\sum_{u=1}^{U}
\mathrm{P@}K_{\mathrm{rec}}^{(m)}(u,\ell),
\qquad
\sigma^{(m)}_\ell
=
\sqrt{
\frac{1}{U}
\sum_{u=1}^{U}
\big(
\mathrm{P@}K_{\mathrm{rec}}^{(m)}(u,\ell)
-
\mu^{(m)}_\ell
\big)^2
}
\end{equation}

        \end{enumerate}

We report Precision@K as the primary metric because the recommendation task in GeMi is explicitly framed as a precision-oriented retrieval problem: given a user's known concept preferences, the system surfaces a short list of K panels likely to match those preferences. In this setting, returning a small number of highly relevant items is more valuable than exhaustive retrieval — a user browsing an art collection benefits more from a curated short list than from high recall across hundreds of panels. We acknowledge, however, that Precision@K alone does not capture ranking quality; future work incorporating NDCG or Mean Average Precision would provide a more complete evaluation, particularly as the dataset grows.
\end{itemize}










\subsection{Results}
\FloatBarrier
In this section, we present empirical analysis in the transductive and inductive settings and answer the research questions proposed in Section~\ref{aims}. 
\subsubsection{Performance Comparison (RQ1)}

\paragraph{Transductive experiments}

\begin{table*}[t]
\centering
\small
\setlength{\tabcolsep}{8pt}
\begin{tabular}{lccc}
\toprule
\textbf{Model} & \textbf{Animal} & \textbf{Mythology} & \textbf{Tree} \\
\midrule
LATTICE  & 0.57 $\pm$ 0.21 & 0.47 $\pm$ 0.25 & 0.20 $\pm$ 0.18 \\
PMGT     & 0.52 $\pm$ 0.21 & 0.46 $\pm$ 0.23 & 0.33 $\pm$ 0.22 \\
HUIGN    & 0.48 $\pm$ 0.16 & 0.55 $\pm$ 0.25 & 0.10 $\pm$ 0.11 \\
DualGNN  & 0.55 $\pm$ 0.21 & 0.47 $\pm$ 0.26 & 0.25 $\pm$ 0.21 \\
FREEDOM  & 0.58 $\pm$ 0.22 & 0.41 $\pm$ 0.24 & 0.22 $\pm$ 0.19 \\
MICRO    & 0.61 $\pm$ 0.20 & 0.47 $\pm$ 0.26 & 0.26 $\pm$ 0.21 \\
DGVAE    & 0.55 $\pm$ 0.21 & 0.45 $\pm$ 0.26 & 0.24 $\pm$ 0.22 \\
\textbf{GeMi GCN} & \textbf{0.47 $\pm$ 0.27} & \textbf{0.61 $\pm$ 0.16} & \textbf{0.62 $\pm$ 0.30} \\
\textbf{GeMi GAE} & \textbf{0.59 $\pm$ 0.30} & \textbf{0.55 $\pm$ 0.13} & \textbf{0.49 $\pm$ 0.29} \\
\textbf{GeMi VGAE} & \textbf{0.51 $\pm$ 0.22} & \textbf{0.69 $\pm$ 0.16} & \textbf{0.48 $\pm$ 0.21} \\
\bottomrule
\end{tabular}
\caption{\textbf{Transductive} Llama-SigCLIP features, homogeneous graph}
\label{tr_llama_sig_hm}
\end{table*}

\begin{table*}[t]
\centering
\small
\setlength{\tabcolsep}{8pt}
\begin{tabular}{lccc}
\toprule
\textbf{Model} & \textbf{Animal} & \textbf{Mythology} & \textbf{Tree} \\
\midrule
LATTICE  & 0.53 $\pm$ 0.22 & 0.54 $\pm$ 0.13 & 0.21 $\pm$ 0.25 \\
PMGT     & 0.47 $\pm$ 0.19 & 0.52 $\pm$ 0.17 & 0.25 $\pm$ 0.20\\
HUIGN    & 0.49 $\pm$ 0.20 & 0.51 $\pm$ 0.12 & 0.23 $\pm$ 0.22 \\
DualGNN  & 0.41 $\pm$ 0.18 & 0.56 $\pm$ 0.32 & 0.19 $\pm$ 0.27 \\
FREEDOM  & 0.51 $\pm$ 0.17 & 0.55 $\pm$ 0.24 & 0.24 $\pm$ 0.16 \\
MICRO    & 0.45 $\pm$ 0.25 & 0.58 $\pm$ 0.23 & 0.26 $\pm$ 0.19 \\
DGVAE    & 0.44 $\pm$ 0.23 & 0.60 $\pm$ 0.21 & 0.27 $\pm$ 0.21 \\
\textbf{GeMi GCN} & \textbf{0.60 $\pm$ 0.24} & \textbf{0.65 $\pm$ 0.20} & \textbf{0.36 $\pm$ 0.26} \\
\textbf{GeMi GAE} & \textbf{0.48 $\pm$ 0.21} & \textbf{0.78 $\pm$ 0.23} & \textbf{0.27 $\pm$ 0.15} \\
\textbf{GeMi VGAE} & \textbf{0.62 $\pm$ 0.25} & \textbf{0.56 $\pm$ 0.19} & \textbf{0.27 $\pm$ 0.28} \\
\bottomrule
\end{tabular}
\caption{\textbf{Transductive} Llama-VAE features, homogeneous graph}
\label{tr_llama_vae_hm}
\end{table*}

\begin{table*}[t]
\centering
\small
\setlength{\tabcolsep}{8pt}
\begin{tabular}{lccc}
\toprule
\textbf{Model} & \textbf{Animal} & \textbf{Mythology} & \textbf{Tree} \\
\midrule
MambaRec     & 0.50 $\pm$ 0.24 & 0.60 $\pm$ 0.17 & 0.12 $\pm$ 0.13\\
PGL  & 0.43 $\pm$ 0.26 & 0.60 $\pm$ 0.20 & 0.25 $\pm$ 0.18 \\
HPMRec    & 0.47 $\pm$ 0.26 & 0.63 $\pm$ 0.21 & 0.34 $\pm$ 0.21 \\
COHESION  & 0.41 $\pm$ 0.23 & 0.59 $\pm$ 0.20 & 0.27 $\pm$ 0.17 \\
SMORE  & 0.40 $\pm$ 0.20 & 0.66 $\pm$ 0.24 & 0.24 $\pm$ 0.16 \\
CMDL    & 0.41 $\pm$ 0.25 & 0.67 $\pm$ 0.22 & 0.23 $\pm$ 0.18 \\
\textbf{GeMi GCN} & \textbf{0.60 $\pm$ 0.24} & \textbf{0.65 $\pm$ 0.20} & \textbf{0.36 $\pm$ 0.26} \\
\textbf{GeMi GAE} & \textbf{0.48 $\pm$ 0.21} & \textbf{0.78 $\pm$ 0.23} & \textbf{0.27 $\pm$ 0.15} \\
\textbf{GeMi VGAE} & \textbf{0.62 $\pm$ 0.25} & \textbf{0.56 $\pm$ 0.19} & \textbf{0.27 $\pm$ 0.28} \\
\bottomrule
\end{tabular}
\caption{\textbf{Transductive} Llama-SigCLIP features (user-item)}
\label{tr_llama_sig_ht}
\end{table*}

\begin{table*}[!h]
\centering
\small
\setlength{\tabcolsep}{8pt}
\begin{tabular}{lccc}
\toprule
\textbf{Model} & \textbf{Animal} & \textbf{Mythology} & \textbf{Tree} \\
\midrule
MambaRec     & 0.52 $\pm$ 0.19 & 0.57 $\pm$ 0.26 & 0.15 $\pm$ 0.28\\
PGL  & 0.44 $\pm$ 0.23 & 0.60 $\pm$ 0.21 & 0.27 $\pm$ 0.21 \\
HPMRec    & 0.46 $\pm$ 0.18 & 0.64 $\pm$ 0.22 & 0.23 $\pm$ 0.14 \\
COHESION  & 0.50 $\pm$ 0.24 & 0.55 $\pm$ 0.11 & 0.25 $\pm$ 0.26 \\
SMORE  & 0.41 $\pm$ 0.54 & 0.59 $\pm$ 0.13 & 0.17 $\pm$ 0.31 \\
CMDL    & 0.57 $\pm$ 0.28 & 0.61 $\pm$ 0.15 & 0.16 $\pm$ 0.25 \\
\textbf{GeMi GCN} & \textbf{0.60 $\pm$ 0.24} & \textbf{0.65 $\pm$ 0.20} & \textbf{0.36 $\pm$ 0.26} \\
\textbf{GeMi GAE} & \textbf{0.48 $\pm$ 0.21} & \textbf{0.78 $\pm$ 0.23} & \textbf{0.27 $\pm$ 0.15} \\
\textbf{GeMi VGAE} & \textbf{0.62 $\pm$ 0.25} & \textbf{0.56 $\pm$ 0.19} & \textbf{0.27 $\pm$ 0.28} \\
\bottomrule
\end{tabular}
\caption{\textbf{Transductive} Llama-VAE features heterogeneous graph}
\label{tr_llama_vae_ht}
\end{table*}


Tables~\ref{tr_llama_sig_hm}--\ref{tr_llama_vae_ht} compare \textsc{GeMi} against state-of-the-art, graph-based baselines under transductive setting across homogeneous item--item graphs and heterogeneous user--item graphs.
Overall, \textsc{GeMi} achieves the strongest and most consistent performance across concepts, particularly on Mythology and Tree, while remaining competitive on Animal.
The results demonstrate that combining LLM-enhanced multimodal features with graph-based representation learning yields clear improvements over prior homogeneous and heterogeneous graph recommenders.

On homogeneous item--item graphs with Llama-SigCLIP features, baselines such as LATTICE, PMGT, HUIGN, DualGNN, FREEDOM, MICRO, and DGVAE achieve moderate performance, with MICRO reaching 0.61 on Animal and HUIGN reaching 0.55 on Mythology.
However, \textsc{GeMi-GCN} substantially improves Tree to 0.62 and Mythology to 0.61, indicating strong relational propagation when label signals are sparse.
\textsc{GeMi-GAE} further improves Animal to 0.59 while maintaining balanced results across labels.
Most notably, \textsc{GeMi-VGAE} achieves 0.69 on Mythology, the highest value in this configuration, showing that variational graph encoding better captures uncertainty and latent structure in item–item relations.
Thus, even when strong homogeneous baselines are considered, all three \textsc{GeMi} variants either match or surpass the best-performing competitors in at least one label dimension.

Under homogeneous graphs with Llama-VAE features, the improvements become more pronounced.
While baselines remain below 0.60 on most dimensions, \textsc{GeMi-GCN} reaches 0.60 on Animal and 0.65 on Mythology.
\textsc{GeMi-GAE} achieves the overall best Mythology performance at 0.78, significantly exceeding prior methods.
\textsc{GeMi-VGAE} reaches 0.62 on Animal, again outperforming baselines.
These results show that when richer generative text representations are used, the graph encoder variants in \textsc{GeMi} consistently dominate homogeneous graph baselines.

A similar trend holds for heterogeneous user--item graphs.
In the Llama-SigCLIP configuration, competitive multimodal CF baselines such as MambaRec, PGL, HPMRec, COHESION, SMORE, and CMDL achieve values around 0.60–0.67 on Mythology but remain weaker on Tree.
\textsc{GeMi-GCN} improves Tree to 0.36 while maintaining 0.65 on Mythology.
\textsc{GeMi-GAE} achieves 0.78 on Mythology, the strongest result in this setting.
\textsc{GeMi-VGAE} reaches 0.62 on Animal, again outperforming heterogeneous competitors.
This indicates that \textsc{GeMi} effectively integrates collaborative user–item structure with multimodal feature signals.

In the Llama-VAE heterogeneous setting, the pattern persists.
Although CMDL and HPMRec perform competitively on Animal and Mythology, none of the baselines reach the 0.78 Mythology score achieved by \textsc{GeMi-GAE}.
Similarly, \textsc{GeMi-GCN} and \textsc{GeMi-VGAE} maintain strong performance on Animal and Tree relative to prior methods.
Across all heterogeneous experiments, the Tree label particularly benefits from the explicit graph propagation and label-aware augmentation in \textsc{GeMi}, where prior CF models struggle due to sparsity.

Collectively, these findings answer RQ1 by demonstrating that \textsc{GeMi}, through its integration of LLM/VLM-derived features and graph-based relational learning, outperforms both homogeneous item--item GNN recommenders and heterogeneous collaborative filtering baselines under transductive training.

\begin{table*}[t]
\centering
\small
\setlength{\tabcolsep}{8pt}
\begin{tabular}{lccc}
\toprule
\textbf{Model} & \textbf{Animal} & \textbf{Mythology} & \textbf{Tree} \\
\midrule
GraphSAGE& 0.48 ± 0.14& 0.56 ± 0.23& 0.34 ± 0.28\\
PinSAGE& 0.52 ± 0.21& 0.57 ± 0.22& 0.49 ± 0.14\\
GATNE-I& 0.52 ± 0.20& 0.60 ± 0.21& 0.37 ± 0.19\\
\textbf{GeMi GCN} & \textbf{0.56 ± 0.24}& \textbf{0.66 ± 0.20}& \textbf{0.44 ± 0.20}\\
\textbf{GeMi GAE} & \textbf{0.50 ± 0.22}& \textbf{0.63 ± 0.15}& \textbf{0.34 ± 0.23}\\
\textbf{GeMi VGAE} & \textbf{0.52 ± 0.22}& \textbf{0.70 ± 0.21}& \textbf{0.44 ± 0.20}\\
\bottomrule 
\end{tabular}
\caption{\textbf{Inductive} Llama-SigCLIP features (item-item)}
\label{homo-sig-ind}
\end{table*}

\begin{table*}[t]
\centering
\small
\setlength{\tabcolsep}{8pt}
\begin{tabular}{lccc}
\toprule
\textbf{Model} & \textbf{Animal} & \textbf{Mythology} & \textbf{Tree} \\
\midrule
GraphSAGE& 0.56 ± 0.25& 0.58 ± 0.12& 0.33 ± 0.16\\
PinSAGE& 0.56 ± 0.25& 0.57 ± 0.12& 0.32 ± 0.17\\
GATNE-I& 0.56 ± 0.25& 0.60 ± 0.19& 0.34 ± 0.22\\
\textbf{GeMi GCN} & \textbf{0.46 ± 0.22}& \textbf{0.32 ± 0.19}& \textbf{0.56 ± 0.34}\\
\textbf{GeMi GAE} & \textbf{0.51 ± 0.29}&\textbf{ 0.61 ± 0.18}& \textbf{0.29 ± 0.23}\\
\textbf{GeMi VGAE} & \textbf{0.50 ± 0.27}& \textbf{0.61 ± 0.23}& \textbf{0.29 ± 0.23}\\
\bottomrule
\end{tabular}
\caption{\textbf{Inductive} Llama-VAE features (item-item)}
\label{homo-vae-ind}
\end{table*}

\begin{table*}[t]
\centering
\small
\setlength{\tabcolsep}{8pt}
\begin{tabular}{lccc}
\toprule
\textbf{Model} & \textbf{Animal} & \textbf{Mythology} & \textbf{Tree} \\
\midrule
GraphSAGE
& 0.63 ± 0.18& 0.59 ± 0.19& 0.26 ± 0.18\\
PinSAGE
& 0.50 ± 0.21& 0.65 ± 0.17& 0.21 ± 0.15\\
GATNE-I& 0.51 ± 0.20& 0.60 ± 0.21& 0.36 ± 0.19\\
\textbf{GeMi GCN} & \textbf{0.71 ± 0.25}& \textbf{0.39 ± 0.05}& \textbf{0.35 ± 0.13}\\
\textbf{GeMi GAE} & \textbf{0.47 ± 0.20}&\textbf{0.71 ± 0.20}& \textbf{0.20 ± 0.22}\\
\textbf{GeMi VGAE} & \textbf{0.52 ± 0.23}& \textbf{0.77 ± 0.11}& \textbf{0.53 ± 0.19}\\
\bottomrule
\end{tabular}
\caption{\textbf{Inductive} Llama-SigCLIP features heterogeneous graph}
\label{heter-sigclip-ind}
\end{table*}

\begin{table*}[t]
\centering
\small
\setlength{\tabcolsep}{8pt}
\begin{tabular}{lccc}
\toprule
\textbf{Model} & \textbf{Animal} & \textbf{Mythology} & \textbf{Tree} \\
\midrule
GraphSAGE
& 0.58 ± 0.24& 0.52 ± 0.15& 0.26 ± 0.12\\
PinSAGE
& 0.46 ± 0.20& 0.61 ± 0.11& 0.39 ± 0.22\\
GATNE-I& 0.56 ± 0.24& 0.61 ± 0.20& 0.33 ± 0.22\\
\textbf{GeMi GCN} & \textbf{0.36 ± 0.12}& \textbf{0.59 ± 0.07}& \textbf{0.35 ± 0.13}\\
\textbf{GeMi GAE} & \textbf{0.69 ± 0.25}&\textbf{0.77 ± 0.11}& \textbf{0.34 ± 0.15}\\
\textbf{GeMi VGAE} & \textbf{0.53 ± 0.19}& \textbf{0.78 ± 0.10}& \textbf{0.53 ± 0.19}\\
\bottomrule
\end{tabular}
\caption{\textbf{Inductive} Llama-VAE features heterogeneous}
\label{heter-vae-ind}
\end{table*}

\paragraph{Inductive experiments}

Tables ~\ref{homo-sig-ind}--\ref{homo-vae-ind} summarize inductive performance under homogeneous item–item graphs. Overall, the best-performing \textsc{GeMi} variants are label-dependent.
\begin{itemize}
    \item Animal: The strongest inductive item–item result is achieved by GeMi-GCN with LLM-SigCLIP features (0.56), which is similar to the best inductive competitors with LLM-VAE. 
    \item Mythology: The best inductive result is GeMi-VGAE with LLM-SigCLIP features (0.70). The strongest competing baseline is GATNE-I (0.60), while GraphSAGE and PinSAGE are lower. 
    \item Tree: Tree is the hardest label overall, but \textsc{GeMi} still achieves the best inductive item–item performance. Under LLM+SigCLIP, GeMi-GCN reaches 0.56 ± 0.34, substantially above GraphSAGE/PinSAGE/GATNE-I. A possible explanation is that ``tree" is often a secondary or background element in many panels: it occupies a small region, is visually less salient than the main subjects, and may be weakly described (if at all) in the accompanying text. As a result, both the multimodal features and the induced neighborhood structure provide a limited discriminative signal for this label compared to more semantically dominant categories such as Mythological characters or Animals. 
\end{itemize}

In the user–item inductive setting (Tables ~\ref{heter-sigclip-ind}--\ref{heter-vae-ind}), \textsc{GeMi} variants remain competitive and sometimes improve further for specific labels (e.g., Animal: GeMi-GCN with LLM-SigCLIP (0.71); Mythology: GeMi-VGAE with LLM-VAE (0.78); Tree: GeMi-VGAE with LLM-VAE  (0.53)), which is expected since the user–item bipartite graph provides additional interaction structure beyond purely feature-based item neighborhoods. 

\subsubsection{Feature Generation (RQ2)}

\noindent \textbf{Transductive experiments}
We first compare base feature representations (SigCLIP + Graph and VAE + Graph in Tables~\ref{tr_base_sig} and \ref{tr_base_vae}) against their LLM-enhanced counterparts within the same GeMi backbone (GCN, GAE, VGAE).
We then analyze the relative behavior of LLM + SigCLIP versus LLM + VAE across label dimensions.

\noindent \textbf{LLM-enhanced vs Base features within GeMi}
Comparing tables~\ref{tr_base_sig} and \ref{tr_base_vae} with tables~\ref{tr_llama_sig_hm}--\ref{tr_llama_vae_ht} results shows a consistent and substantial improvement when LLM-generated features are used.
For example, under homogeneous item--item graphs, Base-SigCLIP + GCN achieves 0.48 on Animal and 0.70 on Mythology, whereas LLM-SigCLIP + GCN improves Animal to 0.60 while maintaining strong Mythology performance.
Similarly, Base-VAE + GCN reaches 0.49 on Animal and 0.38 on Tree, while LLM-VAE + GCN improves Animal to 0.60 and maintains competitive Tree performance at 0.36.

The gains are even more pronounced for the autoencoder variants.
Base-VAE + GAE achieves only 0.33 on Animal and 0.47 on Mythology, whereas LLM-VAE + GAE improves these to 0.48 and 0.78, respectively.
This large margin on Mythology indicates that LLM-enhanced representations better encode semantic structure that benefits reconstruction-based graph objectives.
Likewise, LLM-SigCLIP + VGAE improves Animal from 0.51 (Base) to 0.62 and stabilizes performance across other labels.

Across heterogeneous user--item graphs, the same pattern holds.
While competitive baselines such as MambaRec, HPMRec, SMORE, and CMDL remain within the 0.57--0.67 range on Mythology, the LLM-enhanced GeMi variants reach up to 0.78.
These results confirm that feature quality, rather than solely graph architecture, is a critical factor in performance.

\noindent \textbf{LLM + SigCLIP vs LLM + VAE}
Comparing the two LLM-enhanced feature pipelines reveals complementary behavior across labels.
On Animal, LLM-SigCLIP + VGAE achieves 0.62, slightly outperforming LLM-VAE + GCN at 0.60, suggesting that discriminative contrastive alignment benefits visually grounded categories.
On Mythology, LLM-VAE + GAE achieves the strongest result at 0.78, exceeding LLM-SigCLIP variants, indicating that generative multimodal alignment captures abstract narrative semantics more effectively.
For Tree, LLM-SigCLIP + GCN reaches 0.36, while LLM-VAE + GCN attains a similar 0.36 in heterogeneous settings, showing that both pipelines can capture structural category cues when paired with graph propagation.

Overall, the results demonstrate that LLM-enhanced vision-language representations consistently outperform their base counterparts across all GeMi backbones.
The improvements are particularly significant for semantically rich labels such as Mythology, where LLM conditioning provides clearer contextual signals.

\begin{table*}[t]
\centering
\small
\setlength{\tabcolsep}{8pt}
\begin{tabular}{lccc}
\toprule
\textbf{Model} & \textbf{Animal} & \textbf{Mythology} & \textbf{Tree} \\
\midrule
{SigCLIP + GCN}  & 0.48 $\pm$ 0.25 & 0.70 $\pm$ 0.21 & 0.13 $\pm$ 0.15 \\
{SigCLIP + GAE}  & 0.54 $\pm$ 0.28 & 0.52 $\pm$ 0.17 & 0.20 $\pm$ 0.17 \\
{SigCLIP + VGAE} & 0.51 $\pm$ 0.25 & 0.62 $\pm$ 0.19 & 0.18 $\pm$ 0.18 \\
\bottomrule
\end{tabular}
\caption{\textbf{Transductive} Base-SigCLIP}
\label{tr_base_sig}
\end{table*}

\begin{table*}[t]
\centering
\small
\setlength{\tabcolsep}{8pt}
\begin{tabular}{lccc}
\toprule
\textbf{Model} & \textbf{Animal} & \textbf{Mythology} & \textbf{Tree} \\
\midrule
{VAE + GCN}  & 0.49 $\pm$ 0.23 & 0.58 $\pm$ 0.20 & 0.38 $\pm$ 0.24 \\
{VAE + GAE}  & 0.33 $\pm$ 0.20 & 0.47 $\pm$ 0.21 & 0.23 $\pm$ 0.20 \\
{VAE + VGAE} & 0.34 $\pm$ 0.20 & 0.53 $\pm$ 0.13 & 0.20 $\pm$ 0.18 \\
\bottomrule
\end{tabular}
\caption{\textbf{Transductive} Base-VAE features}
\label{tr_base_vae}
\end{table*}

\noindent \textbf{Inductive experiments}
Following the transductive analysis, we examine whether LLM-enhanced features improve inductive performance within \textsc{GeMi}. Tables ~\ref{ind-sigclip-graph}--\ref{ind-vae-graph} report inductive results using base SigCLIP and base VAE features, while Tables ~\ref{homo-sig-ind}--\ref{heter-vae-ind} report inductive results using LLM-enhanced pipelines (LLM-SigCLIP / LLM-VAE), under the same inductive protocol. In general, LLM-enhanced features outperformed base features, with the magnitude depending on the labels.
\begin{itemize}
    \item Animal: Under the SigCLIP pipeline, LLM-enhancement improves the best \textsc{GeMi} score from 0.52 (Base-SigCLIP + GCN, Table~\ref{ind-sigclip-graph}) to 0.56 (LLM-SigCLIP + GCN, Table~\ref{homo-sig-ind}), suggesting that the normalized text input can strengthen the vision–language embedding used for inductive neighborhood propagation.
    \item Mythology: The best base-SigCLIP result is 0.57 (Table~\ref{ind-sigclip-graph}), whereas LLM-SigCLIP reaches 0.70 (Table~\ref{homo-sig-ind}). This indicates that LLM-enhanced textual message are helpful for mythology, where semantic disambiguation is often text-sensitive.
    \item Tree: Improvements are more mixed. Base-SigCLIP peaks at 0.44 (Table~\ref{ind-sigclip-graph}) and LLM-SigCLIP stays in a similar range (0.44, Table~\ref{homo-sig-ind}), while the VAE pipeline shows a larger jump from 0.51 (Base-VAE + VGAE, Table~\ref{ind-vae-graph}) to 0.56 (LLM-VAE + GCN, Table~\ref{homo-vae-ind}). This is consistent with Tree being a visually subtle category.
\end{itemize}

We further compare the two LLM-enhanced feature pipelines under the inductive setting: LLM-SigCLIP (Table~\ref{homo-sig-ind} and Table~\ref{heter-sigclip-ind}) versus LLM-VAE (Table~\ref{homo-vae-ind} and Table~\ref{heter-vae-ind}). Overall, the better pipeline is label-dependent, and the pattern differs slightly between homogeneous item–item and heterogeneous user–item graphs.
In homogeneous graphs, we observe the following:
\begin{itemize}
    \item Animal: LLM-SigCLIP has the strongest \textsc{GeMi} configuration for Animal (0.56), while LLM-VAE provides comparable baselines but does not surpass the best \textsc{GeMi} score in this label.
    \item Mythology: LLM-SigCLIP is clearly stronger: \textsc{GeMi}-VGAE achieves 0.70, whereas the best LLM-VAE \textsc{GeMi} result is 0.61. 
    \item Tree: The trend reverses: LLM-VAE performs better for Tree under item–item inductive graphs, with \textsc{GeMi}-GCN reaching 0.56  compared to the best LLM-SigCLIP \textsc{GeMi} around 0.44. This indicates that the VAE fusion maybe able to better preserve the background semantics for Tree in the inductive regime.
\end{itemize}
On the other hand, in heterogeneous settings, the two LLM-enhanced pipelines have closer performance. For the animal concept, LLM-SigCLIP attains 0.71 versus 0.69 for LLM-VAE. For the mythology concept, LLM-VAE is slightly higher (0.78) than LLM-SigCLIP (0.77), and for Tree, the best results are essentially tied (0.53 in both user-item Tables). 

Overall, inductive results show that LLM enhancement is consistently beneficial over base features, while the relative advantage between LLM-SigCLIP and LLM-VAE is label- and graph-dependent: SigCLIP tends to favor Mythology in homogeneous graphs, whereas VAE is more helpful for visually subtle labels such as Tree

\begin{table*}[t]
\centering
\small
\setlength{\tabcolsep}{8pt}
\begin{tabular}{lccc}
\toprule
\textbf{Model} & \textbf{Animal} & \textbf{Mythology} & \textbf{Tree} \\
\midrule
{SigCLIP + GCN}  & 0.52 ± 0.26& 0.57 ± 0.14& 0.39 ± 0.27\\
{SigCLIP + GAE}  & 0.50 ± 0.25& 0.54 ± 0.23& 0.37 ± 0.27\\
{SigCLIP + VGAE} & 0.51 ± 0.35& 0.50 ± 0.27& 0.44 ± 0.21\\
\bottomrule
\end{tabular}
\caption{\textbf{Inductive} Base-SigCLIP}
\label{ind-sigclip-graph}
\end{table*}

\begin{table*}[t]
\centering
\small
\setlength{\tabcolsep}{8pt}
\begin{tabular}{lccc}
\toprule
\textbf{Model} & \textbf{Animal} & \textbf{Mythology} & \textbf{Tree} \\
\midrule
{VAE + GCN}  & 0.47 ± 0.27& 0.51 ± 0.28& 0.45 ± 0.23\\
{VAE + GAE}  & 0.39 ± 0.23& 0.61 ± 0.22& 0.30 ± 0.25\\
{VAE + VGAE} & 0.51 ± 0.25& 0.33 ± 0.18& 0.51 ± 0.30\\
\bottomrule
\end{tabular}
\caption{\textbf{Inductive} Base-VAE features}
\label{ind-vae-graph}
\end{table*}

\subsubsection{Graph Structure Learning (RQ3)}
We examine whether transductive or inductive graph training yields better recommendation performance.
We analyze this by comparing GCN, GAE, and VGAE across (i) homogeneous item--item graphs and (ii) heterogeneous user--item graphs, under both training paradigms.

\noindent \textbf{Homogeneous item--item graphs (Llama-SigCLIP)}
Comparing Tables~\ref{tr_llama_sig_hm} (transductive) and the corresponding inductive table shows label-dependent behavior.
For the Animal concept, transductive GAE (0.59) and inductive GCN (0.56) are comparable, but transductive GAE remains slightly stronger.
For the Mythology concept, inductive VGAE (0.70) marginally surpasses transductive VGAE (0.69), indicating that for semantically rich categories, inductive learning can generalize well when graph structure is stable.
However, for the Tree concept, transductive GCN (0.62) clearly outperforms inductive GCN (0.44), suggesting that dense propagation over the full graph during training benefits structurally sparse labels.

\noindent \textbf{Homogeneous item--item graphs (Llama-VAE)}
Under VAE features (Table~\ref{tr_llama_vae_hm}), transductive training is consistently stronger.
For the Animal concept, transductive VGAE (0.62) exceeds inductive variants (around 0.50).
For Mythology, transductive GAE achieves 0.78, substantially higher than inductive GAE/VGAE (around 0.61).
For Tree, transductive GCN (0.36) and inductive GCN (0.56) show complementary strengths, but the inductive advantage is concentrated in GCN only, while GAE and VGAE favor transductive settings.

\noindent \textbf{Heterogeneous user--item graphs}
For Llama-SigCLIP features (Table~\ref{tr_llama_sig_ht}), transductive GeMi variants consistently outperform inductive baselines.
Transductive GCN reaches 0.60 (Animal), 0.65 (Mythology), and 0.36 (Tree), whereas inductive GraphSAGE, PinSAGE, and GATNE-I remain below these values on at least two labels.
The same pattern holds for Llama-VAE features (Table~\ref{tr_llama_vae_ht}), where transductive GAE achieves 0.78 on Mythology, clearly exceeding inductive counterparts (maximum around 0.61).

\noindent \textbf{Model-wise trends (GCN vs GAE vs VGAE)}
Across settings, GCN benefits most from transductive propagation on structurally oriented labels such as Tree.
GAE and VGAE show strong advantages on Mythology under transductive training, particularly with LLM-enhanced features, where reconstruction-based objectives can exploit full-graph connectivity.
Inductive training occasionally matches or slightly exceeds transductive performance on specific labels (e.g., Mythology with VGAE in homogeneous SigCLIP), but these gains are not consistently observed across all labels or graph types.

Overall, transductive training provides more reliable and consistently superior performance, especially in heterogeneous user--item graphs and for semantically complex labels such as Mythology.
Inductive training can generalize competitively in certain homogeneous settings, but it does not consistently outperform transductive learning.
Thus, for the recommendation scenario studied here, transductive graph structure learning yields stronger and more stable results across graph formulations and GeMi variants.

\section{Discussion}
\label{disc}
\begin{itemize}
    \item In Section~\ref{LLM}, we presented an LLM-based semantic text canonicalization procedure. This has been implemented using TinyLlama (\cite{Zhang_24ll}) in the GeMi system. Other LLM variants which can replace TinyLlama include BLOOM (\cite{bloom_23a}), Gemma2 (\cite{Gemma2_24a}), Qwen (\cite{Qwen_25a}) and other open source variants. Experimentation with these pretrained models to study their impact on performance is an active area of research.  
    \item Instead of adapting pretrained language models to downstream tasks, a trend in recent years has been to use the \emph{pretrain, prompt and inference} paradigm, reformulating downstream recommendation tasks with hard/soft prompts (\cite{Liu_23g}). In this paradigm, fine-tuning can be avoided, and the pretrained model itself can be directly employed to predict panel ratings, or even output sub-tasks related to recommendation targets such as explanations. For example, Geng et al. \cite{Geng_23r} have utilized a pretrained CLIP component to convert images into tokens. These tokens are added to textual tokens of an item to create a personalized multimodal soft prompt. Different prompting strategies such as Fixed-pretrained Models, Tuning-free prompting or Prompt and pretrained Tuning can be explored in the GeMi system.
    \item In the current version of the GeMi system, we have not explored the possibility of content creation using diffusion models (\cite{Wei_25p, Lin_25y}). However, given that the multimodal data is not always well aligned (for example, the text of the song may exist but there is no illustration available) it may be possible to use generative AI techniques for content creation with the possibility of enhanced personalization of the recommendation system.    
    \item Our current online system -- UPFAR (described in Appendix~\ref{onlineTool}) can be used to generate data about the users of the system and also learn their preferences. We have incorporated this real-world data into GeMi.
    \item Investigation of algorithms for dynamic pricing mechanisms for narrative scrolls is likely to benefit the marginalized artists who often suffer from severe economic hardships. This will be addressed in a future version of GeMi.
\end{itemize}

\section{Conclusion}
\label{conc}
Singing painters are a group of itinerant storytellers who describe moral and mythological stories in long, narrative, scroll paintings. This performing art is now endangered, with only several hundred painters still practicing the art form. An effort is being made to conserve the artwork and lyrics of their songs for future generations, first by extensive field surveys and data collection, followed by careful annotation and processing of the data leading to the development of a recommendation system -- GeMi. This multimodal, graph-based recommendation system has comparable performance with state-of-the-art transductive and inductive graph learning algorithms and incorporates user preferences as in collaborative filtering. The GeMi system has great potential to be extended with generative AI capabilities, or prompt-based recommendations, and with the incorporation of pricing models to enable the purchase of narrative scrolls from an online platform. 
\backmatter





\bmhead{Acknowledgments}

The first author acknowledges support from the following sources: Gnamm Award from the Asia Research Institute, a faculty grant from the Office of International Education at SUNY, Buffalo, and a fellowship from the National Endowment for the Humanities and the American Institute of Indian Studies. The authors would like to thank Daricha Foundation for its help with data collection during COVID, Jishnu Basak for editing images, and students in a graduate class on ``Optimization Methods in Machine Learning" and ``Distributed Computing and Big Data Technologies" for help with processing data collected from field research and annotations. Finally, this work derives inspiration from many discussions with Elise Auerbach, James Nye, and colleagues from the Digital India Learning Initiatives at the University of Chicago.




\noindent

\bigskip





\begin{appendices}

\section{GNN-based Recommendation Systems}
\label{GnnReco}
GNNs have been extensively used to analyze graph-structured data for node classification (\cite{Kipf_17a, Zhu_21}), link prediction (\cite{Zhang_18a}), and information retrieval tasks (\cite{Zhang_21c}). In recent years, their use in recommendation systems (\cite{Wu_22a, Gao_23, Anand_24a, Sharma_24a}) has become prevalent. A large-scale recommendation system typically has three stages: (a) Matching - filtering candidate items from a very large (usually in millions) set of items, reducing the scale and complexity of the problem; (b) Ranking - Scoring candidate items by a ranking model, and (c) Re-ranking - Updating the ranked list to satisfy additional criteria or business needs. GNNs\footnote{A comprehensive review of GNNs and their use is outside the scope of this work, but we refer the interested reader to the following surveys: (\cite{Wu_22a, Gao_23, Anand_24a, Sharma_24a})} are known to have been used for all three stages. 

\subsubsection{Use of GNNs for Matching}
In the matching stage, GNN-based models are often used to develop sophisticated embedding representations of architectures (\cite{Zhang_21c, Gao_23, Wang_19c, He_20a, Wu_21d, Sun_20p, Ying_18p, Liu_21a, Liu_23a, Yu_20a, Mu_22s, Liu_23y}). These models can capture higher-order similarity among users and items and also capture structural connectivity. A guiding principle for the development of these models is that users with similar interactions tend to have similar preferences, and these can be elucidated through multiple rounds of information propagation. \cite{Berg_18a} proposes a graph auto-encoder framework based on differentiable message passing on a bipartite, user-item interaction graph representing movie ratings. In Neural Graph Collaborative Filtering (NGCF)(\cite{Wang_19c}), the authors argue that embeddings obtained from pre-defined user-item features do not sufficiently capture the collaborative filtering effect; instead they exploit the user-item graph structure by propagating embeddings on it. \cite{Sun_20p} proposes Neighbor Interaction Aware (NIA)-GCN, which can explicitly model the relational information between neighbor nodes and exploit the heterogeneous nature of the user-item bipartite graph. \cite{Wang_20a} claim that user-item interactions are often modeled in a uniform manner, which fails to model diverse relationships and disentangle user intents in representations. Their proposed model, Disentangled Graph Collaborative Filtering (DGCF), solves this problem by modeling a distribution over intents for each user-item interaction, and iteratively refining the intent-aware interaction graphs and representations. EdGe-wise mOdulation (EGO, \cite{Chen_22j}) fusion enhances user-item interactions by adaptively distilling edge-wise multimodal information and learning to modulate unimodal nodes under the supervision of other modalities. Adaptive Anti-Bottleneck MultiModal Graph Learning Network (\cite{Cai_22a}) designs a collaborative representation learning module and a semantic module for learning representations of users and micro-videos, respectively. Spurious correlations are learned in \cite{Du_22y}.

\cite{Liu_21a} notice that GCN-based recommendation models suffer from over-smoothing, and high-order neighboring users with no common interests of a user can be involved in the user`s embedding learning in the graph convolution operation. Consequently, to rectify this, a novel Interest-aware Message-Passing GCN (IMP-GCN) recommendation model, which performs high-order graph convolution inside subgraphs, is proposed. HUIGN (\cite{huign_22}) exhibits user intents in a hierarchical graph structure.

Behavior-driven Feature Adapter (BeFA \cite{Fan_25m}) reduces information drift and omission among extracted features. Multi-level sElf-supervised learNing for mulTimOdal Recommendation (MENTOR, \cite{Xu_25q}) addresses the label sparsity problem and the modality alignment problem by enhancing features and the GCN. EliMRec (\cite{Liu_22w}) was designed to eliminate single-modal bias. DualGNN (\cite{Wang_23a}) addresses multimodal fusion problems arising from missing modalities. Modality alignment for matching is explored in PretrAining Modality-Disentangled Representations Model (PAMD \cite{Han_22a}). 
In Self-Supervised Graph Learning (SGL) (\cite{Wu_21d}), multiple views of a node are generated to maximize the agreement between different views of the same node compared to that of other nodes. This is assumed to rectify problems arising from higher degree nodes exerting a larger influence on recommendations and noisy interactions. MMGCL (\cite{Yi_22t}) design modality edge dropout and modality masking to generate multiple views of users and items. 

DGMRec (\cite{Kim_25a}) disentangles modality features from an information-based perspective and generates missing modality features by integrating aligned features from other modalities and leveraging user preferences for modalities. Modality alignment is further explored in \cite{Tran_22p}. COHESION (\cite{Xu_25a}) promotes a dual-stage fusion strategy to reduce the impact of irrelevant information, refining all modalities. GUIDER (\cite{Li_25b}) denoises user feedback. MoDiCF (\cite{Li_25a}) rectifies the difficulty in accurately generating missing data due to the limited ability to capture modality distributions.

Hamming Spatial GCN (HS-GCN) (\cite{Liu_23a}) develops a hash framework to model both the first- and the high-order similarities in the Hamming space through the user-item bipartite graphs. \cite{Yu_20a} observe that when dealing with very large graphs, the convolution operation is over-simplified to make it computationally efficient, but this over-simplification impairs its normal function. Therefore, a Low Pass Collaborative Filter (LCF) is designed to remove the noise caused by quantization in the observed data, which also reduces the complexity of graph convolution in an unscathed way. To deal with the large size of modality encoders and the complexity of fusing high dimensional features, \cite{Zhuang_25a} proposes the use of a light-weight Wasserstein Knowledge Distillation framework.

\begin{figure}[h]
    \centering
    \begin{subfigure}[b]{0.48\textwidth}
        \centering
        \includegraphics[width=\textwidth]{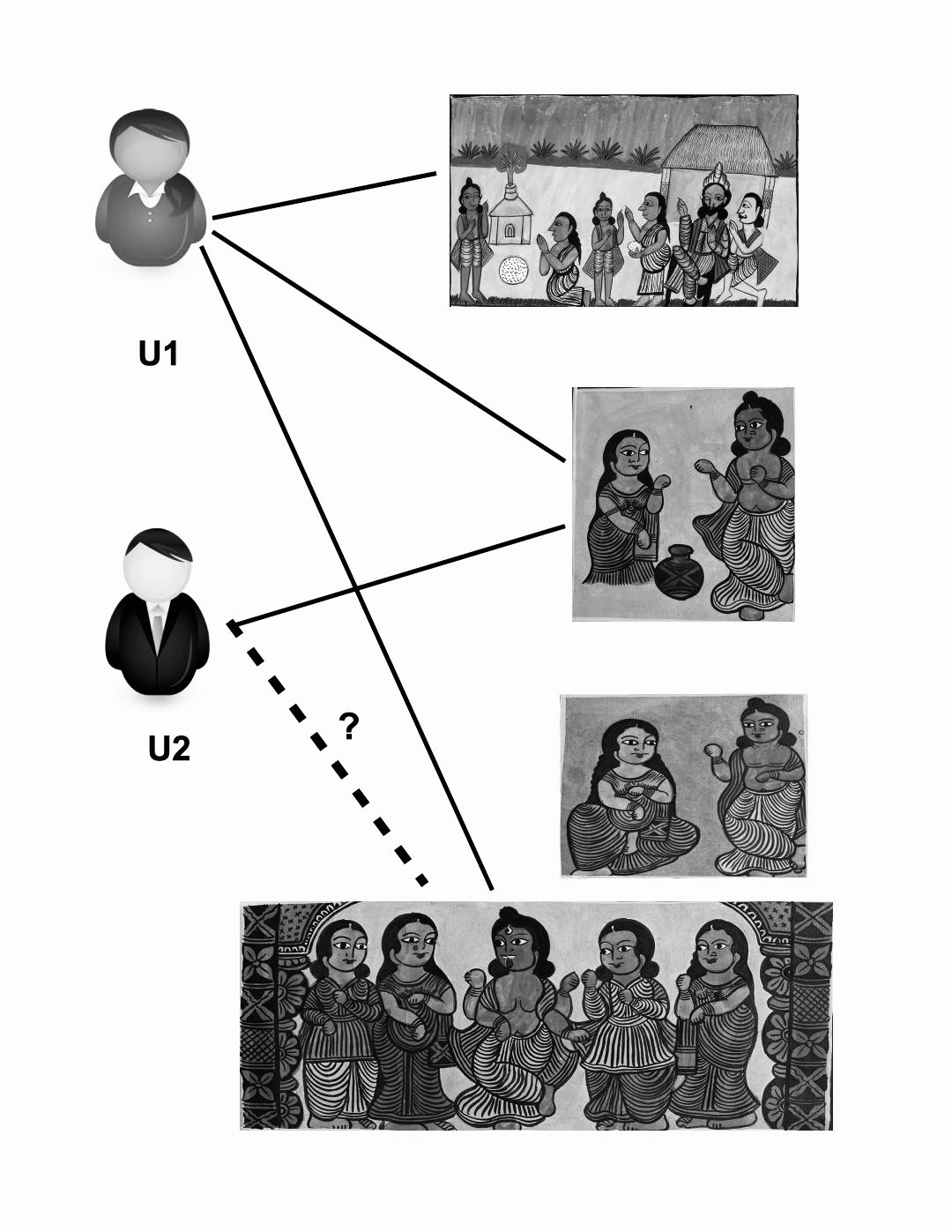}
        \caption{Collaborative item relations}
        \label{fig:sub1}
    \end{subfigure}
    \hfill 
    \begin{subfigure}[b]{0.48\textwidth}
        \centering
        \includegraphics[width=\textwidth]{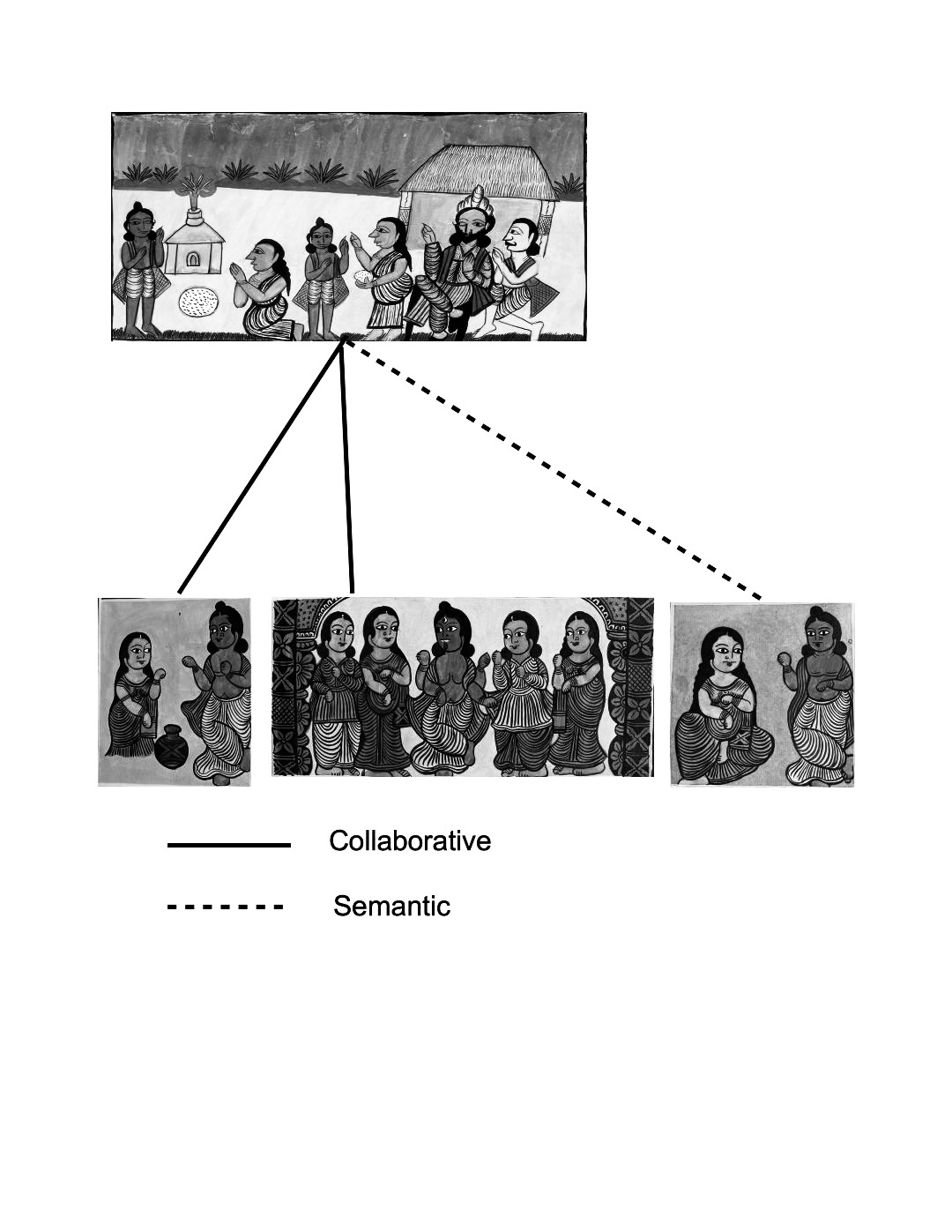}
        \caption{Semantic relationships between items}
        \label{fig:sub2}
    \end{subfigure}
    \caption{A toy example from GeMi illustrating two types of item relations -- collaborative and semantic. Both can be used to learn the latent graph structures.}
    \label{fig:main_figure}
\end{figure}

While GNN models can be used efficiently to inject higher-order connections into embeddings in user-item bipartite graphs, they are known to have several limitations: 
\begin{itemize}
    \item \textbf{Modeling semantic relationships between items:} 
    Figure~\ref{fig:main_figure} illustrates an example from GeMi wherein item-item semantic relationships are as important in understanding latent graph structures as collaborative signals. In the example, panels $\{1,2,4\}$ are of interest to the first user, and $\{2\}$ has only been liked by the second; the collaborative signal helps to learn whether panel $\{4\}$ is of interest. In addition, since the same character is being discussed in all four panels, there is a semantic relation between panel $\{1\}$ and $\{3\}$ which is not inherently captured by this collaborative signal, and the multimodal representation must be able to learn. We are aware of only a few works that capture this semantic information (such as \cite{Meng_25a}).
    \item \textbf{Handling state-of-the-art vision-language models and multimodal fusion:} Many recommendation systems use multimodal information about users and items to model preferences (\cite{Xu_25a} and references there-in), and we are aware of at least two open-source repositories/toolboxes for implementing multimodal system models (MMRec (\cite{zhou_23, Zhou_23a}) and Cornac (\cite{Salah_20, Truong_21a, Truong_21b})). For example, LGMRec (\cite{Guo_24}) first constructs a local graph embedding module to independently learn collaborative and modality-related embeddings of users and items with local topological relations. Then, a global hypergraph embedding module is designed to capture global user and item embeddings by modeling insightful global dependency relations. \cite{pmgt_21} proposes a pretraining strategy to learn item representations by considering both item side information and their relationships. Multimodal Graph Attention Networks (MGAT) (\cite{Tao_20a}) tries to disentangle personal interests at the level of modality. Consequently, multimodal interaction graphs are first built for each modality; following this a gated attention mechanism is used to identify varying importance scores of different modalities to user preference. In the Multimodal Knowledge Graph Attention Network (MKGAT) (\cite{Sun_20a}), a multimodal graph attention technique is used to propagate information on knowledge graphs. In PGMT (\cite{pmgt_21}), the pretrained Inception-v4 network is used to extract image features, and the pretrained BERT model is used to extract sentence features. LATTICE (\cite{Zhang_21a}) learns item-item structures for each modality\footnote{Visual features are obtained from published literature, and textual features embeddings are obtained by concatenating the title, descriptions, categories, and brand of each item utilizing sentence-transformers to obtain a 1,024-dimensional representation.} and aggregates multiple modalities to obtain latent item graphs. To explore highly complicated relationships amongst users a modality-aware Heterogeneous Information Graph (M-HIG) is used to generate high-quality user and micro-video embeddings (\cite{Cai_22}). The model has two key components: (a) a heterogeneous graph that utilizes a random walk-based sampling strategy to sample neighbors for users and micro-videos, and (b) a hierarchical intra- and inter-feature aggregation network to better capture the complex structure and rich semantic information in the heterogeneous graph. 

    Along with GNNs, several algorithms exist which use Graph Auto-Encoders or Variational Graph Auto-encoders (VGAE) for feature extraction. In GraphCAR (\cite{Xu_18a}), two GCNs are used to auto-encode the latent vector space of users and items, and preference scores are obtained by taking the inner product of the two latent factor vectors. In MVGAE (\cite{Yi_22a}), modality-specific variational encoders learn a Gaussian variable for each node -- the mean vector represents semantic information, and the variance vector denotes the noise level of the corresponding modality. In addition, modality-specific Gaussian node embeddings are fused according to the product-of-experts principle, where the semantic information in each modality is weighted based on the estimated uncertainty level. We explored the use of MVGAE for the purpose of ranking in the GeMi system. We use one discriminative (Contrastive Language Image Pretraining (CLIP)) and one generative (VAE) vision language model in the GeMi system (described in detail in Section~\ref{SR}).
    Finally, we are aware of several non-GNN based techniques (such as PAMD) which are not directly relevant to the discussion here. 
    
    \item \textbf{Exploring enhanced linguistic (such as sentence- or discourse-level) properties and controllability (such as prompting techniques) of embeddings using pretrained Large Language Models:} 
    LLMs\footnote{We point the reader to several surveys (\cite{lin_24a, Huang_25a, Wu_24a, Shehmir_25a}) on where and how LLMs can be adapted for use in recommendation systems. This is outside the scope of the current work.} have been used extensively as support tools for recommendation (\cite{Wei_24a, Chen_25z, Ji_25a, Meng_2025p}), such as to develop prompts using user-item data, item metadata, and the candidate items generated by other multimodal recommendation systems. The use of LLMs as supportive components within existing recommendation systems is less explored (\cite{Fioretti_25a}). The GeMi framework explores this supporting role of an LLM to explore sentence and discourse-level linguistic properties of pretrained models (discussed in Section~\ref{LLM}). 
\end{itemize}

\subsubsection{Use of GNNs for ranking}
In the ranking phase of recommendation, a small set of candidate items are identified and a ranking score generated. GNN-based ranking models (\cite{Li_19y, Su_19y, Guo_21y, Li_21y, Wu_22y, Liu_22t}) usually consist of two components - an encoder that captures the desired feature interactions and a decoder (predictor) where the ranking score is estimated by integrating different feature embeddings. \cite{Li_19y} show that simple unstructured combination of feature fields often limits the ability to model sophisticated interactions between features and thus propose to represent the multi-field features in a graph structure, where each node corresponds to a feature field and different fields can interact through edges. \cite{Zheng_23a} study the effect of price in recommendation models by recommending Price-aware user preference modeling (PUP) wherein a heterogeneous graph comprising of four different kinds of nodes -- users, items, prices, and categories is first constructed and embeddings are propagated on them. Finally, a pairwise interaction based decoder is built, leveraging ideas from Factorization Machines (\cite{Rendle_10a}). \cite{Liu_22a} design an Attribute-aware Attentive Graph Convolutional Network ($A^{2}$-GCN) which accounts for missing attributes and allows users to have different preferences for items. \cite{Su_22a} proposes a hypergraph neural network based model, which generates feature interactions of arbitrary order that help learn recommendations. \cite{Guo_21y} design a dual graph enhanced embedding that exploits divide-and-conquer and curriculum-learning-inspired organized learning to address feature sparsity and user-behavior-induced sparsity. \cite{Li_21y} develops a Sequence-aware Heterogeneous graph neural Collaborative Filtering (SHCF) model which incorporates high-order heterogeneous collaborative signals and sequential information to eliminate the data sparsity problem and incorporate dynamic user preferences. 
The GeMi framework explores three different graph architectures -- Graph Convolutional Networks (GCN), Graph Auto Encoders (GAE) and Variational Graph Auto Encoders (VGAE), which are discussed in detail in Section~\ref{GCN}.


\subsubsection{Use of GNNs for Re-ranking}
Re-ranking has been widely accepted as improving recommendation performance on large-scale commercial platforms such as Amazon and Google (\cite{Pei_19a, Liu_20a}). Once candidate items are generated and ranked, a re-ranking algorithm is usually run on top candidates to further refine recommendation results. In \textbf{I}tem \textbf{R}elationship \textbf{G}raph Neural Networks for \textbf{P}ersonalized \textbf{R}ecommendation (IRGPR \cite{Liu_20a}) two graphs - an item relationship graph and an user-item scoring graph are fused to obtain diverse semantic information from heterogeneous node representations. A fully connected GNN is then used to re-rank candidate items. While neural models can be used in the re-ranking phase, it is not uncommon to have simple rule-based functions developed from user preferences or item relationships. In GeMi, we use a label-consistency rule as a re-ranking function.

\FloatBarrier

\begin{table*}[!h]
\centering
\small
\setlength{\tabcolsep}{10pt}
\begin{tabular}{lp{0.78\textwidth}}
\toprule
\multicolumn{1}{c}{\textbf{Model}} & \multicolumn{1}{c}{\textbf{Parameters}} \\
\midrule
LATTICE &
Graph convolution layers: 2;
Graph mixing coefficient ($\lambda$): 0.6;
Loss function: Bayesian Personalized Ranking (BPR);
Optimizer: Adam;
Learning rate: $5 \times 10^{-4}$;
Weight decay: $1 \times 10^{-4}$;
Activation: linear graph propagation;
Embedding normalization: $\ell_2$ normalization;
\\
PMGT &
Neighbors per sequence: 32;
Hidden size: 128;
Transformer layers: 4;
Attention heads: 4;
Feed-forward dimension: 256;
Activation: GELU;
Dropout: 0.1;
Loss functions: BCE-with-logits + MSE;
Masking ratio: 0.16;
Random replacement ratio: 0.02;
Optimizer: AdamW;
Learning rate: $8 \times 10^{-4}$;
Weight decay: $1 \times 10^{-4}$
\\
HUIGN &
Hierarchical intent layers ($K^{(1)},K^{(2)},K^{(3)}$): [32, 8, 4];
Co-interaction threshold: $\ge$ 2;
Loss terms: BPR + entropy regularization + independence regularization + $\ell_2$ weight decay;
Entropy weight ($\lambda_1$): 0.1;
Independence weight ($\lambda_2$): 0.1;
Weight decay ($\lambda_3$): $1 \times 10^{-5}$;
Optimizer: Adam;
Learning rate: $1 \times 10^{-4}$
\\
DualGNN &
Graph propagation depth: 2;
Graph mixing coefficient ($\alpha$): 0.6;
Contrastive temperature ($\tau$): 0.2;
Node feature dropout: 0.2;
Loss terms: InfoNCE alignment + orthogonality regularization + graph smoothness regularization;
Loss weights: $\lambda_{\text{align}}{=}1.0,\,\lambda_{\text{ortho}}{=}0.05,\,\lambda_{\text{smooth}}{=}0.05$;
Optimizer: Adam;
Learning rate: $8 \times 10^{-4}$;
Weight decay: $1 \times 10^{-4}$
\\
FREEDOM &
Multimodal graph mixing: $\alpha_v{=}0.1$ (image) and $\alpha_t{=}0.9$ (text);
Item--Item propagation layers ($L_{ii}$): 1;
User--Item propagation layers ($L_{ui}$): 2;
Degree-sensitive edge pruning ratio ($\rho$): 0.8;
Training loss: BPR + multimodal auxiliary BPR regularizers;
Regularization weight ($\lambda$): $1 \times 10^{-3}$;
Optimizer: Adam;
Learning rate: $1 \times 10^{-3}$;
Weight decay: $1 \times 10^{-5}$
\\
MICRO &
Item-view layers: 1 (per modality);
Graph mixing coefficient ($\alpha$): 0.7;
Contrastive objective: InfoNCE (temperature $\tau{=}0.5$);
Orthogonality regularizer: cross-view residual decorrelation;
Loss weights: $\beta_c{=}0.1$ (contrastive), $\beta_s{=}0.1$;
Optimizer: Adam;
Learning rate: $5 \times 10^{-4}$;
Weight decay: $1 \times 10^{-4}$
\\
DGVAE &
Modality fusion weight ($\alpha_v$): 0.1 (visual), $1-\alpha_v$ (text);
Graph convolution layers: 2;
Number of latent factors ($K$): 4;
Soft assignment temperature ($\tau$): 0.1;
Prior scale ($\sigma_0$): 0.1;
Objective: BCE-with-logits + KL regularization + mutual-information proxy regularization;
Regularization weights: $\beta{=}0.1$ (KL), $\lambda_{\mathrm{MI}}{=}0.2$ (MI);
Optimizer: Adam;
Learning rate: $5 \times 10^{-4}$;
Weight decay: $1 \times 10^{-4}$
\\
\bottomrule
\end{tabular}
\caption{Transductive Homogeneous Item--Item Baseline Parameters}
\label{tab:TransHomoPara}
\end{table*}

\begin{table*}[!h]
\centering
\small
\setlength{\tabcolsep}{10pt}
\begin{tabular}{lp{0.78\textwidth}}
\toprule
\multicolumn{1}{c}{\textbf{Model}} & \multicolumn{1}{c}{\textbf{Parameters}} \\
\midrule
MambaRec &
UI graph layers ($L_{ui}$): 2;
Item--Item layers ($L_{mm}$): 2;
Dimensionality reduction factor: $r=2$;
Normalization + dropout: LayerNorm($d$) + Dropout(0.2);
Loss: BPR pairwise ranking loss + $\ell_2$ regularization + contrastive InfoNCE + optional MMD alignment;
Contrastive weight: 1.0;
InfoNCE temperature: 0.2;
MMD weight: $1\times10^{-2}$;
MMD kernel: RBF with bandwidth $\sigma=1.0$;
Regularization weight: $1\times10^{-3}$;
Optimizer: Adam;
Learning rate: $1\times10^{-3}$;
Weight decay: $1\times10^{-4}$
\\
PGL &
UI graph layers ($L_{ui}$): 2;
Item--Item layers ($L_{mm}$): 2;
Dropout: 0.2;
Loss: BPR pairwise ranking + contrastive InfoNCE loss;
Contrastive temperature ($\tau$): 0.2;
Regularization weight ($\lambda$): $1\times10^{-3}$ on contrastive term;
Optimizer: Adam;
Learning rate: $1\times10^{-3}$;
Weight decay: $1\times10^{-4}$
\\
HPMRec &
Hypercomplex algebra dimension ($A$): 4;
UI graph layers ($L_{ui}$): 2;
Item--Item multimodal layers ($L_{mm}$): 2;
Hypercomplex interaction: Cayley--Dickson multiplication with scale $\alpha=0.1$;
Dropout: 0.2;
Loss: BPR pairwise ranking + $\ell_2$ regularization + self-supervised hypercomplex loss;
Reg weight: $1\times10^{-3}$;
SSL weight: 0.05;
Optimizer: Adam;
Learning rate: $1\times10^{-3}$;
Weight decay: $1\times10^{-4}$
\\
COHESION &
UI graph layers ($L_{ui}$): 2;
Multimodal item--item layers ($L_{mm}$): 2;
UI edge dropout: 0.2;
Post-fusion stabilizer: LayerNorm + Linear + LeakyReLU;
User--User neighborhood aggregation: $k_{uu}=40$;
Loss: BPR pairwise ranking + $\ell_2$ regularization;
Reg weight: $1\times10^{-3}$;
Optimizer: Adam;
Learning rate: $1\times10^{-3}$;
Weight decay: $1\times10^{-4}$
\\
SMORE &
UI graph layers ($L_{ui}$): 2;
Item--Item layers ($L_{ii}$): 2;
Spectrum (FFT) convolution: rFFT over feature dimension with learnable complex filters (3 branches: $w_a,w_b,w_f$);
Nonlinear fusion branch: frequency-domain $(\mathrm{FFT}(z)\odot \mathrm{FFT}(z))\odot w_f$ then iFFT;
Dropout: 0.2;
Auxiliary contrastive loss: InfoNCE;
Contrastive weight: 0.15;
Temperature ($\tau$): 0.2;
Regularization: light $\ell_2$;
Reg weight: $1\times10^{-4}$;
Optimizer: Adam;
Learning rate: $2\times10^{-3}$;
Weight decay: $1\times10^{-4}$;
Gradient clipping: max norm 5.0
\\
CMDL &
UI graph layers ($L_{ui}$): 2;
Content-view augmentation: feature dropout $p=0.25$ + Gaussian noise $\sigma=0.05$;
Item content gating;
Loss: BPR pairwise ranking + $\ell_2$ regularization + contrastive alignment;
Contrastive weight: 0.1;
Contrastive temperatures: $\tau_{pos}=0.2$ (items), $\tau_{neg}=0.2$;
Regularization weight: $1\times10^{-3}$;
Optimizer: AdamW;
Learning rate: $1\times10^{-3}$;
Weight decay: $1\times10^{-4}$
\\
\bottomrule
\end{tabular}
\caption{Transductive Heterogenous User--Item Baseline Parameters}
\label{tab:TransHeteroPara}
\end{table*}

\section{Parameters of Baseline Transductive Algorithms}
\label{secA2}

Tables~\ref{tab:TransHomoPara} and~\ref{tab:TransHeteroPara} present the parameters of the homogeneous and heterogeneous transductive baseline algorithms described in Section~\ref{baseAlgos}.

\begin{table*}[!h]
\centering
\small
\setlength{\tabcolsep}{10pt}
\begin{tabular}{lp{0.78\textwidth}}
\toprule
\multicolumn{1}{c}{\textbf{Model}} & \multicolumn{1}{c}{\textbf{Parameters}} \\
\midrule
GraphSAGE&
Base percentile: $90$,
GNN layers: $2$,
SAGEConv aggregator: max,
SAGEConv settings: normalize: False, 
Hidden channels: $12$, 
Output channels: 2,
Activation: ReLU (between layers),
Loss: cross-entropy on item labels,
Optimizer: Adam,
Learning rate: $0.0001$. \\
PinSAGE&
Percentile threshold: $90$,
Random-walk neighborhood: $50$,
walk length: $5$,
new neighbor per node(topm): $20$,
GNN encoder: same as GraphSAGE above (2 layers, hidden 12, out 2, Adam lr $0.0001$; cross-entropy).\\
GATNE-I&
kNN on training items (k = 30),
Similarity floor: 0.0,
Feature preprocessing: L2-normalize features before cosine,  
Embedding dim: 128,
Negative samples: 10, 
Optimizer: Adam,
Learning rate: $0.001$,
Weight decay: $0.000001$.\\
\bottomrule
\end{tabular}
\caption{Inductive Homogeneous Item--Item Baseline Parameters}
\label{tab:IndHomoPara}
\end{table*}

\begin{table*}[!h]
\centering
\small
\setlength{\tabcolsep}{10pt}
\begin{tabular}{lp{0.78\textwidth}}
\toprule
\multicolumn{1}{c}{\textbf{Model}} & \multicolumn{1}{c}{\textbf{Parameters}} \\
\midrule
GraphSAGE&
GNN layers: 2,
SAGEConv aggregator: max,
SAGEConv settings: project=True, normalize=False,
Hidden channels: 12, 
Output channels: 2,
Activation: ReLU,
Loss: cross-entropy on item labels,
Optimizer: Adam,
Learning rate: 1e-4,
Weight decay: default. \\
PinSAGE&
Percentile threshold: 90,
Random-walk neighborhood: 50,
walk length: 5, 
new neighbor per node(topm): 20,
GNN encoder same as GraphSAGE above (2 layers, hidden 12, out 2, Adam, cross-entropy)\\
GATNE-I&
kNN on training items (k = 30),
Similarity floor: 0.0,
Feature preprocessing: L2-normalize features before cosine,
Embedding dim: 128,
Negative samples: 10,
Optimizer: Adam,
Learning rate: $1e-3$,
Weight decay: $1e-6$.
No learning on test interactions.\\
\bottomrule
\end{tabular}
\caption{Inductive Heterogenous User--Item Baseline Parameters}
\label{tab:IndHeteroPara}
\end{table*}  

\section{Parameters of Baseline Inductive Algorithms}
\label{secA3}
Tables~\ref{tab:IndHomoPara} and~\ref{tab:IndHeteroPara} present the parameters of the homogeneous and heterogeneous inductive baseline algorithms described in Section~\ref{baseAlgos}.

\section{Ablation Studies}
This section presents ablation studies on various parameters of the GeMi system in both transductive and inductive settings. 

\paragraph{Transductive experiments}
We conduct a sensitivity analysis focusing specifically on the shared hyperparameters presented in Table~\ref{tr_ablation}. The objective is to quantify the robustness of GeMi under perturbations of training regularization, graph sparsity, supervised balancing strength, and evaluation configuration. By varying top-$k$ recommendation cutoff, edge dropout rate, focal loss parameters, while keeping architecture-specific components fixed, we isolate the contribution of these global design choices.

\begin{figure*}[t]
    \centering
    \includegraphics[width=1\linewidth, height=0.25\textheight]{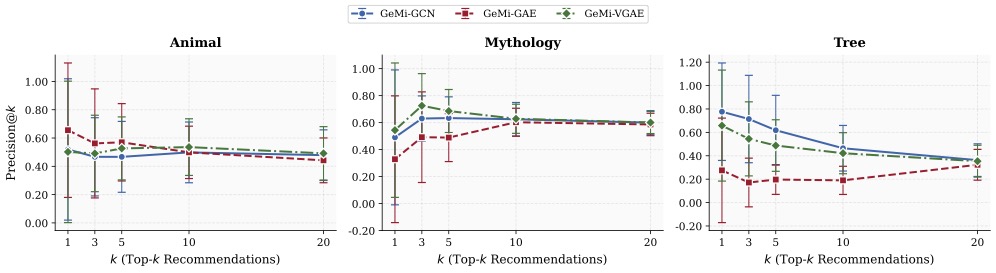}
    
    \vspace{0.5em}
    
    \includegraphics[width=1\linewidth, height=0.25\textheight]{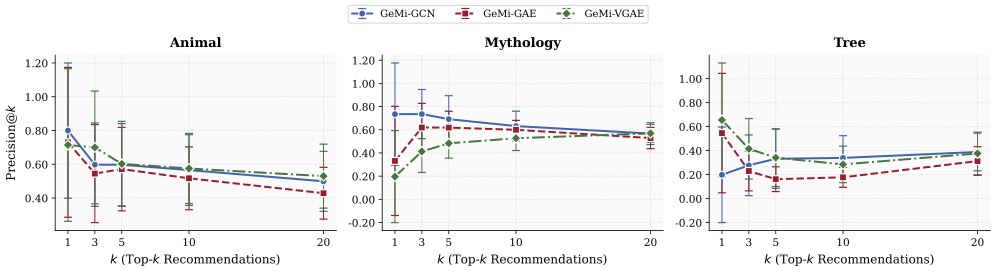}
    
    \caption{Effect of the Top-$k$ recommendation cutoff on Precision@5 across Animal, Mythology, and Tree categories for GeMi variants under different feature backbones (LlamaSigCLIP and LlamaVAE). Results are reported as mean $\pm$ standard deviation over multiple runs.}
    \label{fig:rq4_knn_sensitivity}
\end{figure*}

We analyze the sensitivity of GeMi with respect to the Top-$k$ recommendation cutoff in the transductive setting (Fig.~\ref{fig:rq4_knn_sensitivity}). Across all variants, moderate values of $k$ yield the most stable and reliable performance trends, while very small or large $k$ lead to suboptimal evaluation behavior.

For Animal, Precision@5 is highest at lower $k$ (approximately $k=1$–$5$) and gradually declines as $k$ increases, indicating that the most relevant items are ranked early, while larger recommendation lists introduce less relevant candidates, reducing precision. GeMi-GCN achieves the strongest peak performance, followed by GeMi-VGAE, whereas GeMi-GAE shows comparatively larger degradation as $k$ increases.

For Mythology, performance remains relatively stable across increasing $k$, with slight improvements at moderate values ($k=5$–$10$) before marginal decline. This suggests that semantically abstract categories benefit from a slightly broader recommendation list, though larger $k$ introduces diminishing returns due to inclusion of lower-relevance items. GeMi-VGAE demonstrates greater robustness across the entire range of $k$.

For Tree, which is comparatively sparse and challenging, Precision@5 is lower at very small $k$ and improves with moderate values ($k=5$–$10$), after which it stabilizes. This indicates that a slightly larger candidate pool helps retrieve relevant items, but further increases introduce noise. GeMi-VGAE again exhibits more stable behavior, while GeMi-GAE is the most sensitive.

Overall, the results highlight a trade-off governed by the recommendation cutoff: smaller $k$ emphasizes high-confidence predictions, while larger $k$ increases coverage at the expense of precision. GeMi exhibits controlled sensitivity to this parameter, with consistent performance trends across variants. GeMi-GCN achieves the strongest peak performance, whereas GeMi-VGAE provides greater robustness across varying $k$, supporting the use of moderate Top-$k$ values for balanced and effective recommendation performance.

\begin{figure*}[t]
    \centering
    \includegraphics[width=0.9\linewidth, height=0.17\textheight]{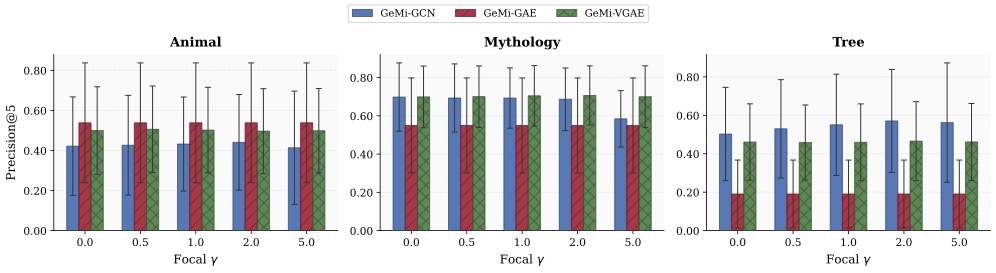}
    
    \vspace{0.5em}
    
    \includegraphics[width=0.9\linewidth, height=0.17\textheight]{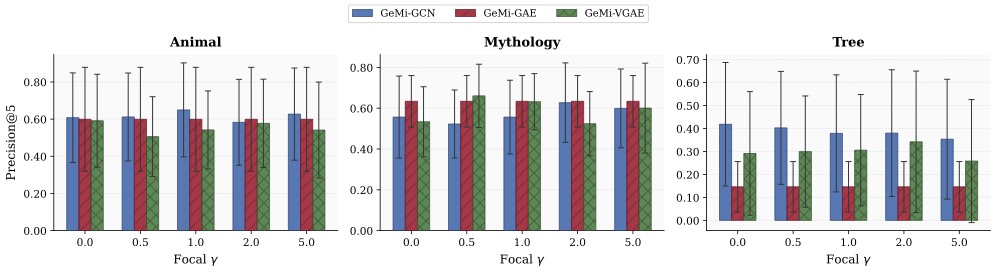}
    
    \caption{Effect of the focal loss focusing parameter $\gamma$ on Precision@5 across Animal, Mythology, and Tree categories for GeMi variants under different feature backbones (LlamaSigCLIP and LlamaVAE). Results are reported as mean $\pm$ standard deviation over multiple runs.}
    \label{fig:rq4_gamma_sensitivity}
\end{figure*}

We further analyze the sensitivity of GeMi with respect to the focal loss focusing parameter $\gamma$, using both LlamaSigCLIP and LlamaVAE feature backbones, as illustrated in Fig.~\ref{fig:rq4_gamma_sensitivity}. Overall, the models exhibit relatively stable behavior across a broad range of $\gamma$ values, indicating that GeMi is not overly sensitive to moderate variations in the focal modulation strength.

For the LlamaSigCLIP setting, the Animal category shows marginal variation across $\gamma \in [0,5]$, with GeMi-GAE and GeMi-VGAE maintaining stable performance, while GeMi-GCN remains slightly lower but consistent. This suggests that for visually grounded and relatively well-separated categories, the benefit of increasing $\gamma$ is limited, as class imbalance is already sufficiently handled by the graph structure and feature quality. In contrast, for Mythology, a slight improvement is observed as $\gamma$ increases from $0$ to around $1$--$2$, after which performance saturates. This indicates that moderate focusing helps emphasize harder samples in semantically abstract categories, but excessive focusing yields diminishing returns. For Tree, performance remains relatively stable with small gains at moderate $\gamma$, suggesting that focal reweighting provides mild benefits for sparse categories but does not drastically alter performance trends.

For the LlamaVAE setting, a similar but slightly more pronounced trend is observed. In the Animal category, GeMi-GCN benefits from moderate $\gamma$ (around $1$), achieving peak performance, while higher values introduce slight instability, likely due to over-emphasis on difficult samples. GeMi-GAE and GeMi-VGAE remain relatively stable, though GeMi-VGAE shows minor degradation at extreme $\gamma$ values. In the Mythology category, both GeMi-GAE and GeMi-VGAE achieve improved performance at moderate $\gamma$ (around $0.5$--$1$), reflecting better handling of hard and ambiguous samples, while very high $\gamma$ leads to marginal decline due to over-focusing. For the Tree category, which is the most imbalanced, moderate $\gamma$ improves performance for GeMi-VGAE, but GeMi-GCN shows slight degradation as $\gamma$ increases, indicating that excessive focusing may suppress useful majority-class signals.

Across both feature settings, the results highlight that $\gamma$ controls a trade-off between emphasizing hard samples and maintaining stable gradient contributions from easier examples. Lower values ($\gamma \approx 0$) reduce focal loss to standard BCE, leading to more uniform learning, while higher values ($\gamma > 2$) over-amplify difficult samples, potentially introducing optimization instability. Empirically, moderate values ($\gamma \in [0.5, 2]$) consistently yield the best or near-optimal performance across categories and model variants.

In summary, GeMi demonstrates robustness to the focal loss parameter, with performance varying smoothly across $\gamma$. Moderate focusing provides consistent benefits, particularly for semantically complex and imbalanced categories, while extreme values are unnecessary and can slightly degrade performance. These findings support the choice of $\gamma=2.0$ as a balanced and effective default across different feature backbones and GeMi variants.

\begin{figure*}[t]
    \centering
    \includegraphics[width=0.9\linewidth, height=0.19\textheight]{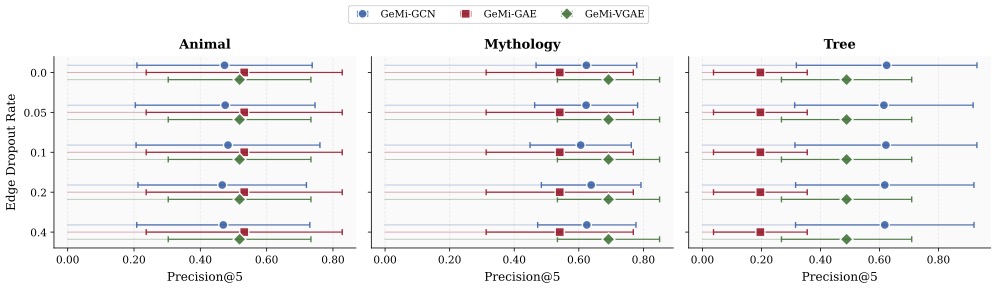}
    
    \vspace{0.5em}
    
    \includegraphics[width=0.9\linewidth, height=0.19\textheight]{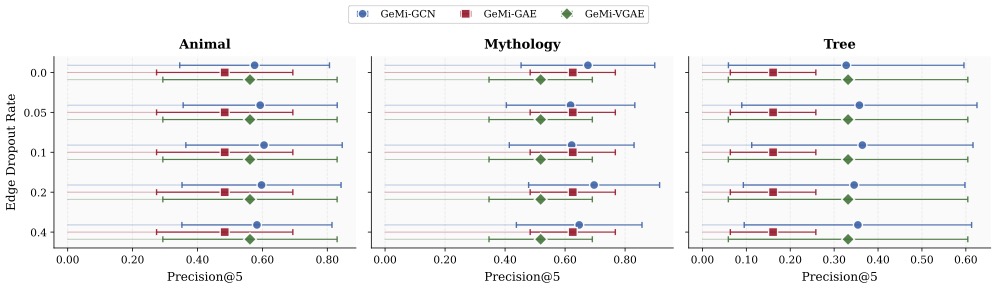}
    
    \caption{Effect of the edge dropout parameter on Precision@5 across Animal, Mythology, and Tree categories for GeMi variants under different feature backbones (LlamaSigCLIP and LlamaVAE). Results are reported as mean $\pm$ standard deviation over multiple runs.}
    \label{fig:rq4_edgedrop_sensitivity}
\end{figure*}

We finally analyze the sensitivity of GeMi to edge dropout, which induces stochastic sparsification of the interaction graph during training, under both LlamaSigCLIP and LlamaVAE feature backbones (Fig.~\ref{fig:rq4_edgedrop_sensitivity}). Overall, GeMi demonstrates strong robustness, with only minor variations in Precision@5 across dropout rates from $0.0$ to $0.4$, indicating stable representation learning under structural perturbations.

In the LlamaSigCLIP setting, performance remains largely consistent across all dropout levels. For Animal, GeMi-GCN shows a slight improvement at moderate dropout ($\sim 0.1$), reflecting mild regularization benefits, while higher dropout slightly degrades performance due to excessive edge removal. GeMi-GAE and GeMi-VGAE remain nearly invariant. In Mythology, GeMi-VGAE consistently achieves the best performance ($\sim 0.69$) with negligible variation, while GeMi-GCN exhibits mild gains at moderate dropout. In Tree, GeMi-GCN and GeMi-VGAE remain stable, indicating that moderate sparsification does not hinder information propagation even in sparse regimes.

In the LlamaVAE setting, slightly higher sensitivity is observed. For Animal, GeMi-GCN benefits from moderate dropout (peak around $0.1$), suggesting improved generalization via removal of redundant edges, while higher dropout reduces performance. In Mythology, GeMi-GCN peaks at moderate dropout ($\sim 0.2$), indicating effective noise filtering, whereas excessive sparsification degrades performance. GeMi-GAE remains stable, and GeMi-VGAE shows consistent but lower performance. In Tree, moderate dropout again benefits GeMi-GCN, while higher dropout reduces connectivity and slightly harms performance.

Overall, the results highlight a trade-off between regularization and information loss. Low dropout preserves full connectivity but retains noise, moderate dropout ($0.05$--$0.2$) improves robustness and generalization, and high dropout ($\geq 0.4$) weakens message passing. GeMi-VGAE exhibits the highest robustness, while GeMi-GCN benefits most from moderate regularization. These findings support the use of moderate edge dropout as an effective and stable design choice.

\paragraph{Inductive experiments}
We study the sensitivity of inductive performance to the kNN graph parameter $k$, which controls (i) the density of the training item-item subgraph and (ii) the connection of each test item to the training set. Across labels and backbones, the effect of $k$ is non-monotonic: increasing $k$ is not uniformly beneficial, and in some cases smaller $k$ will generate better performance. This result implies a bias-variance tradeoff where larger $k$ would introduces more weakly similar neighbors and may causing the result over smoothing, while too small $k$ can lead to an overly sparse graph where information passing become weaker.

\textbf{Non-monotonicity: }The impact of $k$ can be strongly non-monotonic even when holding the model and feature pipeline fixed. For instance, under a SigCLIP + GAE setting, performance improves markedly when decreasing $k$ from 30 to 10 ,
but drops again at $k=5$. This implies that the best fit knn for graph density lies in a middle range rather than at extremes.

\textbf{Label-dependent sensitivity: } The optimal $k$ is label-dependent, but the sensitivity pattern is not uniform across backbones and feature pipelines. In many settings, Tree label tends to be more k-sensitive, while Mythology can range from relatively stable to strongly $k$-sensitive. A reasonable explanation is that Tree is a weaker signal (often visually subtle and not always dominant in text), so the choice of neighborhood size more directly affects whether sufficient informative signal can be aggregated, additionally, this also motivates our Tree-specific edge augmentation.

\textbf{Feature dependence: }The preferred $k$ is also feature dependent. Different embedding leads to different nearest-neighbor quality. A smaller $k$ can preserve higher-quality neighbors in one embedding space but may overly sparsify the graph in another. Therefore, $k$ acts as a smoothing hyperparameter for message passing, and its preferred range depends on the base feature geometry.

\textbf{Backbone dependence: }Compared with GCN and GAE, VGAE tends to show higher sensitivity to $k$, with performance sometimes shows sharper rises and drops as $k$ varies. This aligns with VGAE's generative objective, where changes in graph density can reshape the latent structure and lead to local optima.

\begin{figure}
    \centering
    \includegraphics[width=1\linewidth]{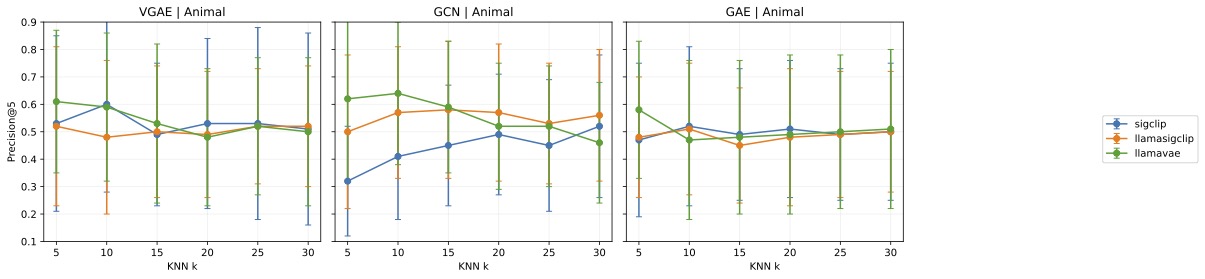}
    \caption{RQ4 (kNN ablation): Precision@5 vs. kNN graph construction parameter \textit{k} for Animal. Error bars denote mean ± 1 std across runs. }
    \label{fig:Animal}
\end{figure}
\begin{figure}
    \centering
    \includegraphics[width=1\linewidth]{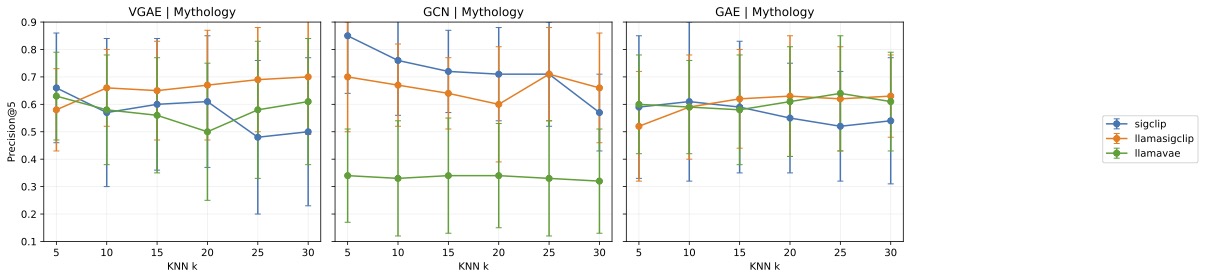}
    \caption{RQ4 (kNN ablation): Precision@5 vs. kNN graph construction parameter \textit{k} for Mythology. Error bars denote mean ± 1 std across runs.}
    \label{fig:Mythl}
\end{figure}

\begin{figure}
    \centering
    \includegraphics[width=1\linewidth]{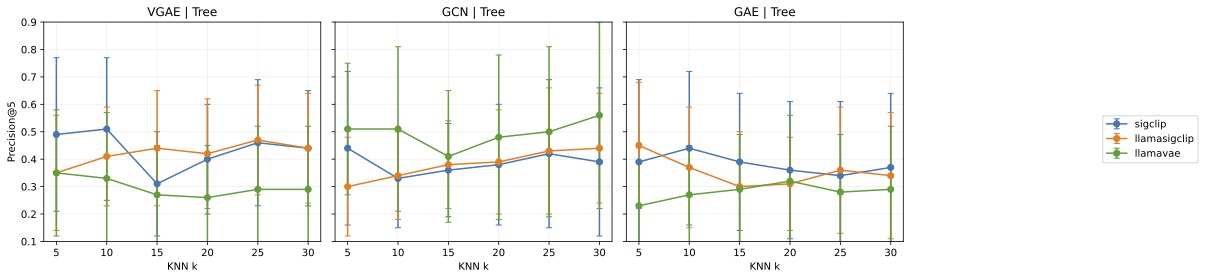}
    \caption{RQ4 (kNN ablation): Precision@5 vs. kNN graph construction parameter \textit{k} for Tree. Error bars denote mean ± 1 std across runs.}
    \label{fig:Treel}
\end{figure}

\section{An online tool for folk art recommendation}
\label{onlineTool}
\begin{figure}[!t]
    \centering
    \includegraphics[width=0.75\linewidth]{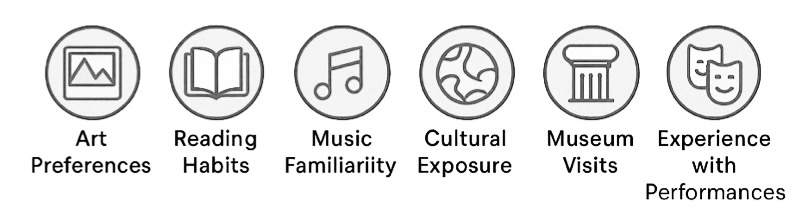}
    \caption{Characteristics of users working with UPFAR }
    \label{fig:Char}
\end{figure}

\begin{figure}
    \centering
    \includegraphics[width=0.75\linewidth]{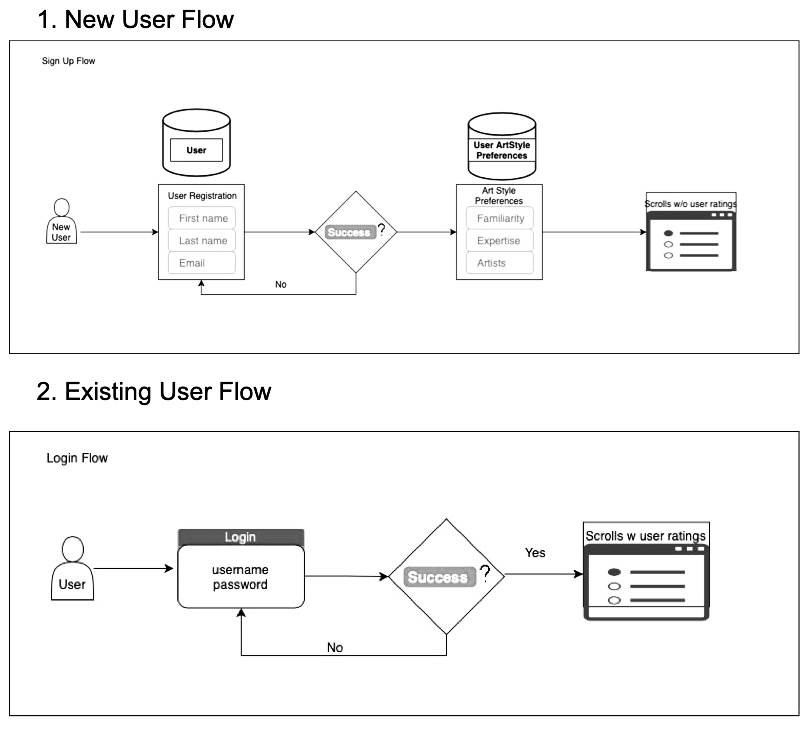}
    \caption{Architecture diagram of UPFAR}
    \label{fig:archUPFAR}
\end{figure}

\begin{figure}
    \centering
    \includegraphics[width=0.95\linewidth]{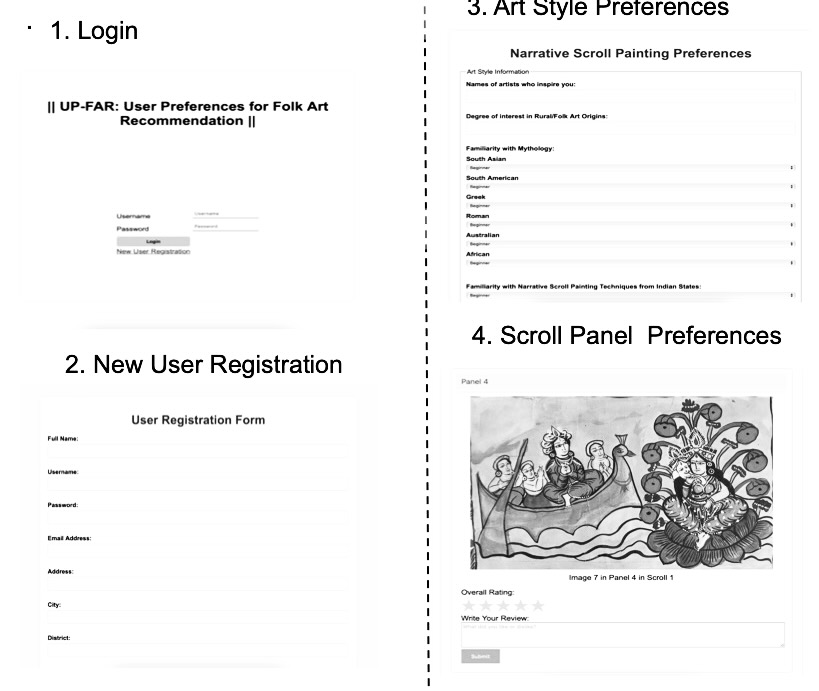}
    \caption{Deployed screens from UPFAR}
    \label{fig:deploy}
\end{figure}

In ongoing work (\cite{Hotwani_25a}), the authors have developed a system -- User Preferences for Folk Art Recommendation (UPFAR) -- to collect information (such as cultural exposure, museum visits, art and music familiarity, experience with prior performances of narrative scroll paintings) and ratings (like($=1$)/not($=0$)) for panels. Figure~\ref{fig:Char} shows the categories of data collected from the users, Figure~\ref{fig:archUPFAR} the architecture diagram, and the tool deployed online (Figure~\ref{fig:deploy}). Data from this system was ingested into the GeMi pipeline. The following details are relevant:
\begin{itemize}
\item Data pre-processing: The user-panel interaction data was cleaned by removing invalid entries, removing duplicates, and keeping only those interactions that belong to the training panels, to avoid any leakage of test information into user profiles.
\item Computing preferences: To convert raw ratings into user preference over labels, the ratings were normalized using min-max normalization in $[0,1]$. Then, 
\begin{itemize}
    \item For each user, a baseline rating (their average normalized rating) was computed.
    \item For each label (animal, myth, tree), the average rating of items that contain that label was computed.
\end{itemize}
Using this, relative lift is defined as follows: 
\begin{equation*}
lift = \text{average rating for that label} - \text{user baseline}
\end{equation*}
which implies that if a user rates a concept higher than their average, it gets a positive lift, otherwise negative. 
\item Dealing with sparsity: Many users have very few interactions for a label. So, a Bayesian-style smoothing was applied that combined the user’s lift with a global prior lift (computed across all users). This is weighted by how many examples the user has, thus preventing unstable estimates for sparse users.
\item Developing binary preferences: The lift for each user is mapped into a probability-like preference using a sigmoid function as follows:
\begin{equation*}
\text{preference} = \sigma(\text{gain} \times \text{lift}). 
\end{equation*}
The gain controls how sharply the preferences move away from $0.5$. This gives us a clean preference matrix, where each user has a value between $0$ and $1$ for animal, myth, and tree concepts. At the same time, for each user, we also store their top-k interacted panels $(k=5)$ based on highest ratings. This is used later to build user embeddings during evaluation.
\item Augmentation of user profiles: The real-world data collected from UPFAR is very small. To circumvent this problem, we have generated a large, realistic dataset through bootstrapped augmentation. For each synthetic user (target = 10,000 users), we randomly pick a real user as a base, sample 5 interactions from that user’s top-k items (with replacement), and with some probability $(p=0.3)$, we randomly replace a few items with other observed items to introduce diversity. For preferences, the base user’s base preference vector is used to apply gain scaling (to make some users more extreme or more neutral), small bias shift (to simulate user tendency) and gaussian noise per label (to diversify), and then clip back values to the desired range $[0,1]$.
\end{itemize}
Table~\ref{trans-real-data} presents results from the transductive setting. Our results indicate that graph-structure learning using VGAE has significant benefits for this data, but either SIGCLIP or VAE can be used to learn the concepts. Interestingly, only little benefit is obtained from enhancements via LLM canonicalization (for example to learn mythology concepts using SIGCLIP features). Table~\ref{ind-real-data} presents results from the inductive setting. SIGCLIP with GCN or GAE representations are able to provide good recommendations, while the Tree concept remains hard to recommend for real-world data as well. 

\begin{table*}[t]
\centering
\small
\setlength{\tabcolsep}{8pt}
\begin{tabular}{lccc}
\toprule
\textbf{Model} & \textbf{Animal} & \textbf{Mythology} & \textbf{Tree} \\
\midrule
{SIGCLIP + GCN}  & $0.52 \pm 0.22$ & $0.68 \pm 0.17$ & $0.40 \pm 0.20$\\
{SIGCLIP + GAE}  & $0.55 \pm 0.22$ & $0.49\pm0.14$ & $0.33\pm0.22$ \\
{SIGCLIP + VGAE} & $0.44\pm0.16$ & \textcolor{red}{$0.74\pm0.18$} & $0.43\pm0.22$ \\
{VAE + GCN} & $0.27 \pm 0.09$ & $0.59 \pm 0.04$ & $0.01 \pm 0.04$ \\
{VAE + GAE} & $0.50\pm0.22$& $0.69\pm0.11$ & $0.22\pm0.16$ \\
{VAE + VGAE} & \textbf{\textcolor{red}{$0.60\pm0.11$}}& $0.65\pm0.09$ & \textcolor{red}{$0.60\pm0.21$} \\
{Llama SIGCLIP + GCN} & $0.56 \pm 0.19$ & $0.71 \pm 0.16$ & $0.44\pm 0.25$ \\
{Llama SIGCLIP + GAE} & $0.43\pm0.20$ & $0.58\pm0.15$ & $0.40\pm0.17$ \\
{Llama SIGCLIP + VGAE} & $0.59\pm0.16$ & $0.69\pm0.15$ & $0.44\pm0.13$\\
{Llama VAE + GCN} & $0.45 \pm 0.19$ & $0.53 \pm 0.19$ & $0.39 \pm 0.24$ \\
{Llama VAE + GAE} & $0.43\pm0.20$ & $0.43\pm0.23$ & $0.33\pm0.16$ \\
{Llama VAE + VGAE} & $0.48\pm0.17$ & $0.39\pm0.20$ & $0.31\pm0.22$ \\
\bottomrule
\end{tabular}
\caption{Results from \textbf{Transductive} model training using real-world data to guide user-preference matrix generation. }
\label{trans-real-data}
\end{table*}

\begin{table*}[t]
\centering
\small
\setlength{\tabcolsep}{8pt}
\begin{tabular}{lccc}
\toprule
\textbf{Model} & \textbf{Animal} & \textbf{Mythology} & \textbf{Tree} \\
\midrule
{SIGCLIP + GCN}  & $0.45 \pm 0.14$ & \textcolor{red}{$0.79 \pm 0.25$} & $0.23\pm0.23$\\
{SIGCLIP + GAE}  & $0.39\pm0.20$ & $0.58\pm0.20$ & $0.24\pm0.26$ \\
{SIGCLIP + VGAE} & \textcolor{red}{$0.65\pm0.27$} & $0.71\pm0.13$ & $0.30\pm0.21$ \\
{VAE + GCN} & $0.64\pm0.18$ & $0.61\pm 0.09$ & $0.24\pm0.30$ \\
{VAE + GAE} & $0.53\pm0.09$& $0.54\pm0.13$ & $0.19\pm0.16$ \\
{VAE + VGAE} & $0.60\pm0.20$ & $0.64\pm0.19$ & $0.25\pm0.19$ \\
{Llama SIGCLIP + GCN} & $0.56 \pm 0.27$ & $0.52 \pm 0.19$ & $0.25\pm 0.20$ \\
{Llama SIGCLIP + GAE} & $0.49\pm0.22$ & $0.54\pm0.13$ & $0.36\pm0.23$ \\
{Llama SIGCLIP + VGAE} & $0.47\pm0.21$ & $0.59\pm0.17$ & $0.26\pm0.14$\\
{Llama VAE + GCN} & $0.31\pm0.23$ & $0.31\pm0.15$ & $0.33\pm0.17$ \\
{Llama VAE + GAE} & $0.51\pm0.18$ & $0.56\pm0.12$ & \textcolor{red}{$0.38\pm0.12$} \\
{Llama VAE + VGAE} & $0.53\pm0.19$ & $0.48\pm0.13$ & $0.37\pm0.17$ \\
\bottomrule
\end{tabular}
\caption{Results from \textbf{Inductive} model training using real-world data to guide user-preference matrix generation. }
\label{ind-real-data}
\end{table*}




\end{appendices}

\bibliography{sn-bibliography}

\end{document}